\documentclass[9pt,twocolumn,twoside]{osajnl_patched}
\journal{ol}
\renewcommand*{\journalname}{Science Robotics -- Morgan \textit{et al.}}
\setboolean{shortarticle}{false}
\usepackage{fix-cm}   
\usepackage{graphicx}

\usepackage{url}

\newcommand{\athenazero}{\textsc{AthenaZero}}

\definecolor{editcolor}{RGB}{0, 0, 0}
\definecolor{editcolor2}{RGB}{0, 0, 0}
\definecolor{editcolor3}{RGB}{0, 0, 0}

\def\scititle{
	\athenazero{}: A low-inertia, bimanual \\robot for dynamic manipulation
}
\title{\bfseries \boldmath \scititle}

\author[1$^\ast$]{Andrew S. Morgan}
\author[1]{Gregory Xie}
\author[1]{Capprin Bass}
\author[1]{Rachel Thomasson}
\author[1]{Chunpeng Wang}
\author[1]{Erfan Shahriari}
\author[1]{Harrison Busa}
\author[1]{Oluwaseun Araromi}
\author[1]{Joseph Aronov}
\author[1]{Michael Burgess}
\author[1]{Velin D. Dimitrov}
\author[1]{Matthew A. Estrada}
\author[1]{Faris Hamdi}
\author[1]{Samuel Kendig}
\author[1]{Taeyoon Lee }
\author[1]{\\Jose Oscar Mur-Miranda}
\author[1]{Emma Sommers}
\author[1]{Paul Titchener}
\author[1]{Margaret Wang}
\author[1]{Achu Wilson}
\author[1]{Mark Yeatman}
\author[1]{Osman Dogan Yirmibesoglu}
\author[1$^\dagger$]{Alfred A. Rizzi}
\author[1$^\dagger$]{Annan Mozeika}
\author[1$^\dagger$]{\\Nicolas Rojas}
\author[1$^\dagger$]{Lael Odhner}

\affil[1]{RAI Institute, Cambridge, Massachusetts, 02142, USA}
\affil[$^\ast$]{Corresponding author. Email: \texttt{andy@rai-inst.com}}
\affil[$^\dagger$]{Co-Principal Investigators.}

\begin{abstract}

\textcolor{editcolor}{\athenazero{} is a bimanual manipulator designed to minimize inertia without compromising control authority. By utilizing quasi-direct drive actuation and transmission remotization techniques, the system achieves an effective endpoint mass comparable to that of a human---about an order of magnitude less than conventional robot manipulators. This characteristic, combined with its inherent torque transparency, makes \athenazero{} exceptionally well-suited for dynamic manipulation. We describe the methodology} that led to this design and demonstrate the robot's capabilities on three baseball-inspired tasks: throwing, catching, and batting, which showcase complex interactions on human-comparable timescales where milliseconds matter. \athenazero{} was capable of throwing at speeds in excess of 30~m/s, with catching and batting at speeds in excess of 14~m/s over a short 7.3~m distance. Batting practice and a game of catch were subsequently performed in robot-to-robot and human-to-robot variations, showcasing the efficacy and adaptability of our system in tasks that require high acceleration.

\end{abstract}

\begin{document}

\maketitle


\noindent
{\color{editcolor3}\textbf{Article Summary:} \athenazero{} is a bimanual robot designed to minimize inertia and maximize control authority for dynamic manipulation.}

\label{sec:intro}

{\color{editcolor3}\section*{Introduction}}
{\color{editcolor}
\noindent The development of custom robot platforms, albeit arduous, is an indispensable component for progressing the field of robotics. Foundational research of this nature necessitates a substantial investment of time and a commitment to long-term, iterative improvements. One need only look at the last twenty years of progress in locomotion to find evidence for this. The proliferation of increasingly capable walking robots, which has been enabled by advancements in simulation, control, and electric actuator design, has led to a revolution in robotic running, climbing, and acrobatics. {\color{editcolor3}Although} robotic manipulation has seen a parallel explosion in algorithmic capability---particularly in learning-based control and perception---hardware development has lagged, leaving researchers largely reliant on a selection of commercial platforms designed decades ago. Looking forward, mechanical paradigms must evolve to support these rapid algorithmic strides. We posit that low-inertia architectures will {\color{editcolor3} substantially} expand the manipulator's dynamic range, enabling manipulation to extend into dynamics rather than just quasi-statics.

Similar to dynamic walking, a ``dynamic" approach to manipulation does not necessarily imply ``high speeds," but rather that the robot's inertial and torque properties are matched to the tasks it attempts to perform. In locomotion, the dynamic scaling problem is often posed in terms of the natural tipping frequency of the robot. In other words, if a robot can reposition its feet faster than it falls, it remains controllable. Due to the robot's low mass relative to the ground, uncertainty in contact and impact force can be managed.
The dynamic scaling problem in manipulation differs structurally from locomotion. Manipulation involves interacting with environments whose dynamics are not predetermined, often involving comparatively lightweight objects. As a result, the robot's effective mass typically dominates the interaction at the moment of impact. If a mass mismatch occurs, the robot transmits excessively large contact forces, causing the object to accelerate faster than the manipulator and ejecting it. The ability for a system to expose only appropriately scaled interaction forces, and therefore avoid such unintended motion either actively or passively, is at the crux of the robot manipulation problem.

Overcoming contact uncertainty has long motivated manipulation research, with approaches ranging from compliant control to variable impedance actuators \cite{hogan1985,buerger_hogan_CRC,hirzinger2002_lwr,pratt1995series,vanderborght2013variable}. While this prior work has increasingly improved safety and interaction performance, these advancements have primarily targeted large-scale behaviors or specific, task-driven requirements. A robot’s capacity for rapid reactivity to external stimuli or disturbances is a fundamental prerequisite for managing environmental uncertainty.

To produce a robot that is more agile, particularly in contact, high control authority---or the ability to quickly and deliberately accelerate across a wide band of joint velocities---is paramount. We believe that well-behaved contact interactions require leveraging actuation to not just produce fast, reactive motions, but also to dynamically adapt the interaction impedance behavior during contact. The inertia of a system plays a vital role in this capability, in that higher inertia negatively affects control authority if actuator power and torque bandwidth are finite. A grounding metric for intuitive evaluation of a system's inertia---or resistance to a change in motion---is effective mass (formally defined in the following section). Intuitively, this metric tells you how much mass is ``felt" when interacting with the robot at different points on the physical system. As detailed in the Results section, our analysis reveals a {\color{editcolor3}substantial} disparity in effective mass between humans and leading collaborative robots. We posit that the ideal effective endpoint mass for high-bandwidth dynamic manipulation is considerably lower than that of current commercial arms. Rather, it is likely closer to that of a human (2.76 kg, as reported in \cite{kim2017_lims1})---representing an order-of-magnitude reduction compared to most research platforms. This observation serves as the foundation for our robot manipulator design and system-level analysis.


\noindent This article is driven by a central premise:

\begin{quote}
\textit{A robot design that thoughtfully balances maximizing control authority with minimizing effective mass, while supporting realistic payloads ($\ge$3~kg), can yield higher-bandwidth dynamic performance across a wide manipulation range; from delicate, impedance-shaped interactions to powerful, energy-transfer motions.}
\end{quote}

Rooted in decades of learnings, spanning from early direct drive arms \cite{asada1987direct} to contemporary quasi-direct drive (QDD) walking robots \cite{seok2012_cheetah1}, we seek to evaluate if recent advancements in computation and motor technology make a QDD paradigm viable for realizing the above thesis in a bimanual manipulator.  Specifically, we investigate whether a balance can be struck between achieving human-like effective mass and retaining the strength and stiffness necessary for dynamic tasks, all without relying on complex transmissions that compromise system maintainability. We approach this through a combination of custom QDD actuators and mass remotization.

As a result, \athenazero{} achieves an effective mass on the order of that of a human, and control authority substantially higher than conventional robots, enabling dynamic operation across a wide range of scenarios---whether in free space or in contact. As historical efforts have revealed, these benefits come with tradeoffs. By eschewing high-ratio gear reductions ($\geq$80:1) for lower inertia, the system sacrifices static holding torque limits and passive mechanical stiffness compared to its counterparts. However, contemporary computation providing low latency feedback can help alleviate this characteristic through active modulation as compared to decades ago. The following sections detail how we navigated tradeoffs in design and control, assess to what extent we can realistically reduce effective mass, and finally, demonstrate the ability of \athenazero{} by performing tasks that require both agility and precision: throwing, catching, and batting (Fig.~\ref{fig:splash}, {\color{editcolor3} Movie M1).

\begin{figure}[!t]
\centering
\includegraphics[width=\columnwidth,height=0.45\textheight,keepaspectratio]{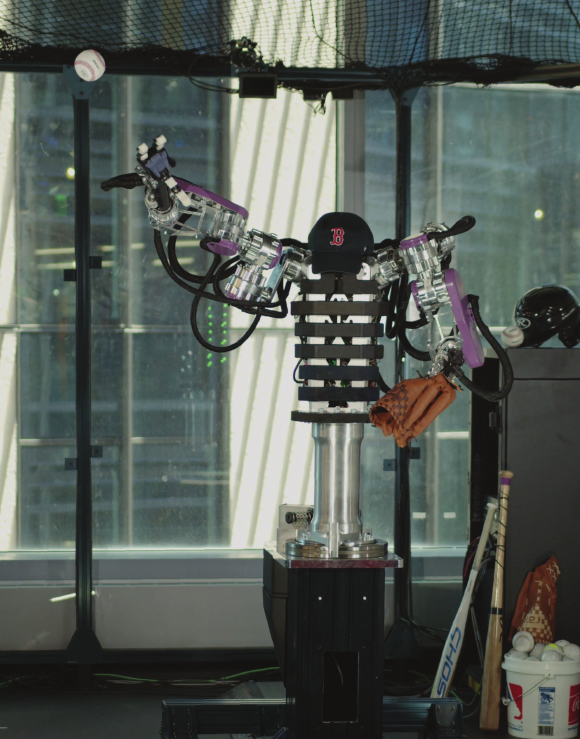}
\caption{{\color{editcolor3}\textbf{\athenazero{} throwing a baseball.} \athenazero{}'s low-inertia design lends itself to highly dynamic, athletic tasks such as throwing a baseball in excess of 25~m/s. }}
\label{fig:splash}
\end{figure}

\subsection*{Preliminaries}
We will commonly reference three closely related but distinct concepts---actuator reflected inertia, manipulator effective mass, and system transparency---which serve as guiding principles to our design methodology.

\textbf{Actuator reflected inertia} captures the apparent inertia of an actuator as seen at the load. When a motor drives a load through a gear reduction of $N$:$1$,  the velocity and torque are scaled by $N$ from the perspective of the load, but the motor’s rotor inertia is amplified by $N^{2}$. Likewise, any damping or stiffness of the rotor---whether from friction or the control loop---is also amplified by $N^{2}$.

\textbf{Effective mass} represents the apparent mass of a manipulator as reflected through its kinematics, and is an operational- or task-space quantity that depends on the configuration of the robot. At an arbitrary point on the robot, the effective mass relates applied wrenches at that point to the resulting accelerations. Link inertia, link mass, and joint reflected inertia all contribute to the effective mass of the robot.

\textbf{Transparency} corresponds to the efficient and bidirectional transmission of force and velocity between the actuators and the operational point(s) of the robot, allowing the actuators to effectively ``feel" the interaction. This property is closely related to backdrivability, since both describe how efficiently external forces pass through the transmission. Factors such as friction, backlash, and unmanaged cogging reduce the transparency of a system.

\subsection*{Background and Challenges}
{\color{editcolor}
Decades of work in manipulator design have repeatedly revisited fundamental design trade-offs as application demands evolved. Early industrial manipulators prioritized precision and repeatability through high structural stiffness and highly geared actuation. Later developments introduced interaction-aware control and mechanically compliant designs to enable safe physical interaction with users and the environment \cite{hogan1985,buerger_hogan_CRC,hirzinger2002_lwr,haddadin2022franka,Clark2024}. To further push the boundaries of force transparency and dynamic performance---enabling highly reactive tasks such as ball catching \cite{bauml2011catching}---researchers have increasingly turned to low-inertia paradigms. Notable examples include the Barrett WAM \cite{townsend1993_wam}, AMBIDEX \cite{kim2017_lims1}, and Blue \cite{gealy2019quasi}.

Despite this apparent range in design approaches, researchers overwhelmingly utilize commercially available ``collaborative robots," or cobots. Cobots are chosen due to their reliability and perceived safety, but despite the implication of their name, their safety stems from programmed safeguards rather than their intrinsic inertial properties. This class of manipulator frequently relies on heavily geared motors ($\geq$80:1) to increase positioning precision and decrease static power consumption. This serves to minimize the influence of environmental dynamics, which improves tracking accuracy but drastically increases reflected inertia (recall that rotor inertia is amplified by $N^2$). Beyond this actuation paradigm, typical embodiments collocate joints (degrees of freedom, or DoF) and actuators for design simplicity. That is, motors are ``stacked" serially to reduce complexity and cost, which in turn increases the moving mass of the manipulator. A combination of high reflected inertia and distal mass equates to a relatively high effective mass, which requires these robots to operate far below their speed limits in contact-rich environments to maintain safety. This issue can be confronted in two ways: hiding the effective mass at contact or physically reducing it.

One way to hide effective mass is through mechanical compliance, such as series elastic actuation \cite{pratt1995_sea, fitzgerald2013_baxter, bicchi2004_fastandslow}, soft or underactuated end-effectors \cite{dollar2010highly, hughes2016_soft}, or link padding \cite{park2011_padding}. Nominally, these modifications are appropriate for specific use-cases, for example fruit picking \cite{navas2021soft} or finger gaiting \cite{morgan2022complex}. However, elastic elements introduce a fundamental tradeoff often in what we care most about:~control authority. Functionally, any compliant element in series with a force source limits the closed-loop bandwidth according to the system's natural frequency. This can often result in a system that is overly ``floppy" and unable to apply forces desirably. Fundamentally, an infinite power actuator coupled to a compliant element with infinite range of travel and accurate state sensing could in theory provide any range of impedances, but this cannot be practically realized.
}

Another way to hide effective mass is by artificially reducing it through force or torque feedback, but fundamental challenges arise. At the theoretical level, Hogan \& Buerger \cite{hogan2018impedance} showed that even for a linear time-invariant mass–spring–damper system, force feedback gain is strictly limited: gains above unity render the system non-passive, leading to negative apparent inertia and potential instability in interaction. Albu-Schäffer \textit{et al.} \cite{albu2007unified} demonstrated that joint torque feedback can scale down motor inertia in flexible joint robots, but only within limits \textcolor{editcolor}{(typically by a factor of about 4–6),} and stability is guaranteed only for quasi-static cases. In principle, \textcolor{editcolor}{arbitrarily} shaping a robot’s effective mass requires not only perfect force/torque sensing but also an accurate full dynamic model \cite{ott2008cartesian} and actuators with infinite bandwidth, since sampling rate, delay, and actuator dynamics impose strict lower bounds on realizable virtual inertia \cite{colgate1997passivity}. \textcolor{editcolor}{In practice, these sensing, modeling, and actuation limitations impose strict bounds on the achievable inertia reduction \cite{dietrich2021practical}, limiting it to moderate levels relative to the robot’s natural inertia. Moreover, virtual inertia reduction is inherently reactive: the apparent inertia can only be modified after contact forces are measured, so the initial impact impulse is governed by the robot’s physical inertia. This limitation is particularly critical for interaction with unknown environments, where reducing the initial impact forces is essential for safe physical contact.}

\textcolor{editcolor}{In contrast to the above approaches that aim to hide effective mass through compliance or feedback, another strategy is to actually reduce the effective mass by directly lowering the apparent moving mass of the manipulator. Prior work has studied its role in interaction dynamics, safety, and task performance \cite{kirschner2025categorizing,steinecker2022mean}.} This approach, albeit more difficult as it requires a direct modification to the robot, can be achieved by reducing link inertia through careful structural optimization with advanced materials \cite{hirzinger2002_lwr}, leveraging suspension elements such as counterweights \cite{wyrobek2008_pr1, whitney2014_counterbalance}, or aggressively remoting actuators through complex transmissions like the Barrett WAM and AMBIDEX \cite{kim2017_lims1, townsend1993_wam, kim2024_axisymmetric, quigley2011low}. Reducing inertia is advantageous as it does not inherently limit control bandwidth while indeed reducing inertia of the manipulator. However, depending on the implementation, such designs can quickly become mechanically complex, for example with intricate tendon routing schemas, which drastically increases cost and decreases reliability of the system as a whole.

Aside from link inertia, the largest contributor to effective mass is often the actuator inertia. Upon first inspection, reducing actuator reflected inertia might seem impossible: idealized scaling laws show that the actuator reflected inertia is invariant to rotor diameter for a given output torque and constant motor mass~\cite{seok2012_cheetah1}. However, in practice, gears introduce additional mass, inertia, and friction.
Using larger, power-dense motors with no or low gear ratios---known as direct drive or quasi-direct drive (QDD) actuators---has the desirable effect of lowering reflected inertia. A QDD joint paradigm can substantially reduce the manipulator's effective mass and move its characteristics closer to those of a human arm \cite{gealy2019quasi}, while also improving force transparency \cite{wensing2017_cheetah}. Their inherent backdrivability, especially with high efficiency ($\geq$97$\%$) planetary gearsets, allows torques to be ``felt" through the transmission. This greatly simplifies actuator design since the motor driver can rely on $q$-axis current, or the quadrature axis current known as the ``torque producing current" in field-oriented control, to estimate torque transmitted through the gearbox and thus, joint-level strain gauge-based torque sensors are not required with proper characterization. Gearbox efficiency, in turn, is critical and works favorably since low gearing typically also results in lower friction and backlash. Overall, relying on transparency and eliminating strain gauge-based output torque sensors yields a simpler system, free from drift and hysteresis that can be exacerbated by heat and impacts.

\subsection*{Design  \textcolor{editcolor}{Methodology}}
\label{sec:design_philosophy}
{\color{editcolor}
Balancing competing trade-offs and system requirements is inherent to robot design. In this study, we intentionally relax thermal and power constraints to focus on minimizing inertia and replicating human-like peak torque \cite{mayer1994normal, kotte2018normative, yoshii2015measurement}. This yields payload performance comparable to a human, albeit with lower static holding capacity than cobots that prioritize rigidity, such as the Franka FR3 \cite{franka}, KUKA iiwa \cite{KUKAi-iwa}, and Universal Robots UR series \cite{universalrobots}.

{\color{editcolor2}
By challenging the requirement for indefinite static holds---designing instead for sustained durations on the order of minutes---\athenazero{} leverages recent advancements in actuation and computation to strike a different balance. Combining highly efficient gearsets with targeted stator and rotor properties, our custom quasi-direct drive actuators drastically reduce reflected inertia, thereby lowering the effective mass at the end-effector ({\color{editcolor3}Movie s1}). We further minimize overall system mass by employing simple, stiff transmissions to relocate actuators toward the robot's center of mass, and by skeletonizing rigid links wherever feasible. Furthermore, modern advances in computation and networking have substantially reduced system latency, enabling higher-bandwidth control. According to Colgate's stiffness scaling \cite{colgate1988}, this reduction in time delay allows the system to modulate its impedance over a much broader range than previously attainable, even if it cannot fully replace the intrinsic passive stiffness lost by removing high-ratio gearboxes.}

}
\section*{Results}
\subsection*{The \athenazero{} Robot}
\label{sec:design}

\athenazero{} is a bimanual robot consisting of a 1~DoF torso, two 7~DoF arms, and two 6~DoF hands (Fig. \ref{fig:schematic}). The entire robot is powered by a family of four custom quasi-direct drive, and highly force-transparent actuators in 95~mm, 76~mm, 38~mm, and 25~mm diameter variations. The robot stands at about 1.6~m tall with an approximate wingspan of 1.8~m. The shoulders have a 0.44~m lateral separation, modeled after a human male. The torso joint is connected to an immobile base, and powered by two 95~mm actuators, providing equivalently a 10.48:1 planetary/belt reduction. \athenazero{} relies on $q$-axis current measurements for torque sensing at the motor-level. Thus, there are no force-torque sensors or joint torque sensors on this robot. The robot is powered by an external 48~V power source with a peak current of 300~A.

\begin{figure}[!ht]
\centering
\includegraphics[width=\columnwidth,height=0.6\textheight,keepaspectratio]{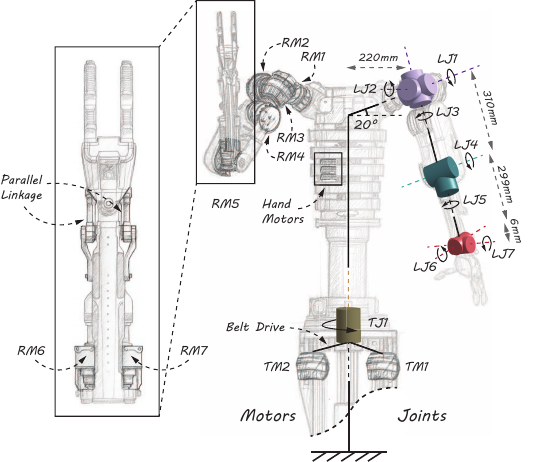}
\caption{\textbf{Motor-level and joint-level architecture of \athenazero{}.}\textit{ Nomenclature: Torso (T), Right (R), Left (L), Joint (J), and Motor (M)}. Two torso motors (TM1 and TM2) connect directly to the TJ1 torso joint in parallel via a belt. Shoulder motors RM1-RM3 (LM1-LM3) are axially coincident, combining towards RJ1-RJ3 (LJ1-LJ3) and acting as a spherical shoulder joint. Elbow junction actuators RM4-RM5 (LM4-LM5) through a belt transmission are axially perpendicular, mimicking a universal joint RJ4-RJ5 (LJ4-LJ5) at the elbow. Wrist motors RM6-RM7 (LM6-LM7) are differentially coupled, symbolically representing a near-universal joint RJ6-RJ7 (LJ6-LJ7) at the wrist junction. Translational offset between the two axes is 6~mm. A total of 3 motors for each hand are located within the torso, with Bowden transmissions that remote actuation out to three fingers via tendons.  }
\label{fig:schematic}
\end{figure}

\begin{table*}
    \small 
    \centering
    \begin{tabular*}{\linewidth}{@{\extracolsep{\fill}} ccccccc }
        Group & Joint & Motor Diameter & Gear Ratio & Peak Torque & Peak Velocity & Actuator Inertia*\\
        \hline
        \hline
        \textit{Torso} &  TJ1 &  2 x 95~mm $\dagger$ & 10.48:1 & 378~Nm & 32~rad/s & 5.49e-2~kg\,m$^2$\\
        \textit{Shoulder} & (R/L)J1/2/3 & 95~mm & 5:1 & 90~Nm & 66~rad/s & 6.25e-3~kg\,m$^2$ \\
        \textit{Elbow} & (R/L)J4 & 95~mm & 5:1 & 90~Nm & 66~rad/s & 6.25e-3~kg\,m$^2$ \\
         & (R/L)J5 & 76~mm & 5:1 & 20~Nm & 89~rad/s & 1.28e-3~kg\,m$^2$\\
        \textit{Wrist} & (R/L)J6 & \textbf{38~mm}\textbar \textbar38~mm $\ddagger$& $\sim$10.6:1 & $\sim$16.1~Nm & $\sim$100~rad/s & 4.09e-4~kg\,m$^2$\ \\
         & (R/L)J7 & 38~mm\textbar \textbar\textbf{38~mm} $\ddagger$& $\sim$7.5:1 & $\sim$11.4~Nm & $\sim$142~rad/s  & 2.28e-4~kg\,m$^2$\
    \end{tabular*}
    \caption{\textbf{\athenazero{} joint specifications.} \\
    $\dagger$ TJ1 is actuated by two 95~mm motors coupled in parallel with a belt reduction.\\
    $\ddagger$ The wrist is a differentially coupled parallel mechanism. Wrist specifications are taken at the neutral position and due to the properties of the linkage, is non-linear.\\
    * Actuator Inertia signifies the total reflected rotor inertia of that motor/gearbox combination.}
    \label{tab:joint_specs}
\end{table*}

\begin{figure*}[!t]
\centering
\includegraphics[width=\textwidth,height=0.26\textheight,keepaspectratio]{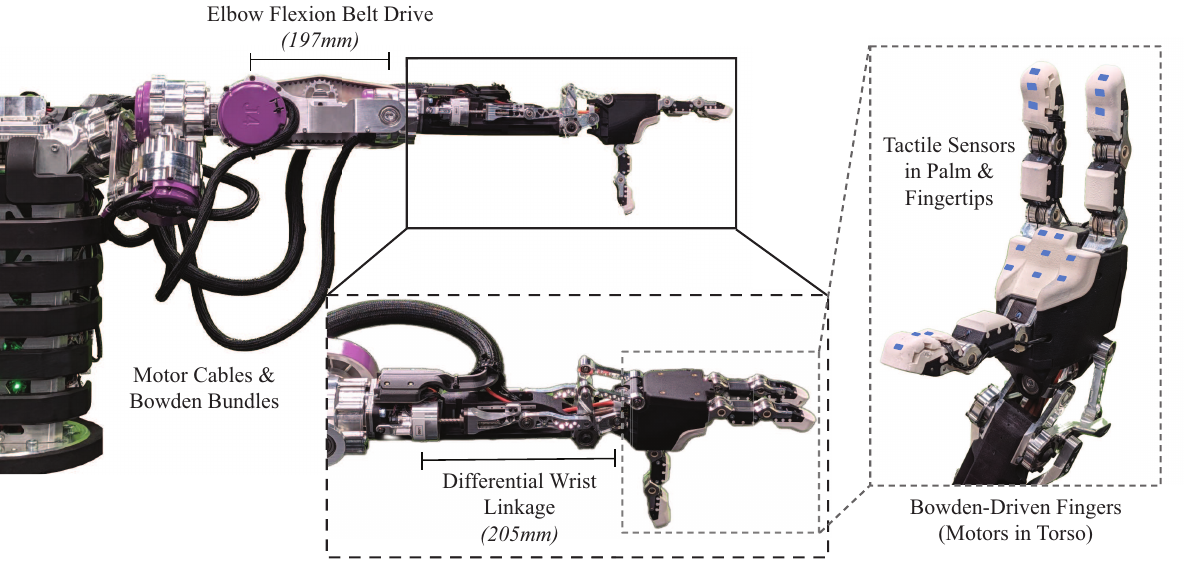}
\caption{\textbf{Remoting actuation via transmissions reduces effective mass.} \textit{Shoulder and Elbow: }The majority of the arm's mass is contained via the combination of motors RM1-RM5 (equivalently LM1-LM5). Shoulder motors are located at the joint, but elbow flexion leverages a belt transmission. \textit{Wrist: } The wrist is a 2~DoF parallel mechanism which remotes the motors to the elbow junction. \textit{Hand: }The hand is a 6~DoF, 3-tendon underactuated mechanism that interdigitates. Tactile sensors are placed inside of the fingertips and palm for contact sensing (denoted in blue). }
\label{fig:shoulder_wrist_hand}
\end{figure*}

Requirements for intuitive teleoperation and human-like kinematics led us to choose a quasi-anthropomorphic 3~DoF-2~DoF-[1~DoF-1~DoF] joint paradigm for the shoulder, elbow, and wrist junctions, respectively (Fig.~\ref{fig:shoulder_wrist_hand}). The wrist, notably, is not coaxial but has a small 6~mm offset. Joint specifications are provided in Table \ref{tab:joint_specs}. Given this mechanical architecture, we evaluated morphologies such as parallel shoulder mechanisms and differential drive joints, and transmission methods such as tendons, belts, hydraulics, and pneumatics. Several design criteria for reducing reflected inertia resulted in a hybrid serial-parallel design. This design struck a desirable balance between modularity and durability. The hands, further described in the Methods section, are underactuated and were optimized towards grasping a baseball and a baseball bat. Full control of \athenazero{}, including the hands, requires coordination of 22 actuators over a total of 27 joints.

\subsection*{Effective Mass Analysis}
\label{sec:effective_mass_analysis}
{\color{editcolor}
We compared the effective mass of \athenazero{} to four leading commercial robots. The Franka Research 3 (FR3) \cite{franka}, Universal Robots UR5e \cite{universalrobots}, and KUKA LBR iiwa 14 \cite{KUKAi-iwa} utilize high gear reductions ($\geq$80:1), whereas the 7~DoF Barrett WAM~\cite{townsend1993_wam} uses QDD actuation similar to \athenazero{}. Compared to \athenazero{}, the WAM features more complex transmissions to remote its heaviest actuators to the base. Fig.~\ref{fig:impedance}A compares the effective mass of these robots and a representative human arm~\cite{NASA3001Vol2}, evaluating contact at the wrist in a neutral configuration. For each robot, we reported values both with and without reflected rotor inertia.

\begin{figure*}[!p]
\centering
\includegraphics[width=\textwidth,height=0.5\textheight,keepaspectratio]{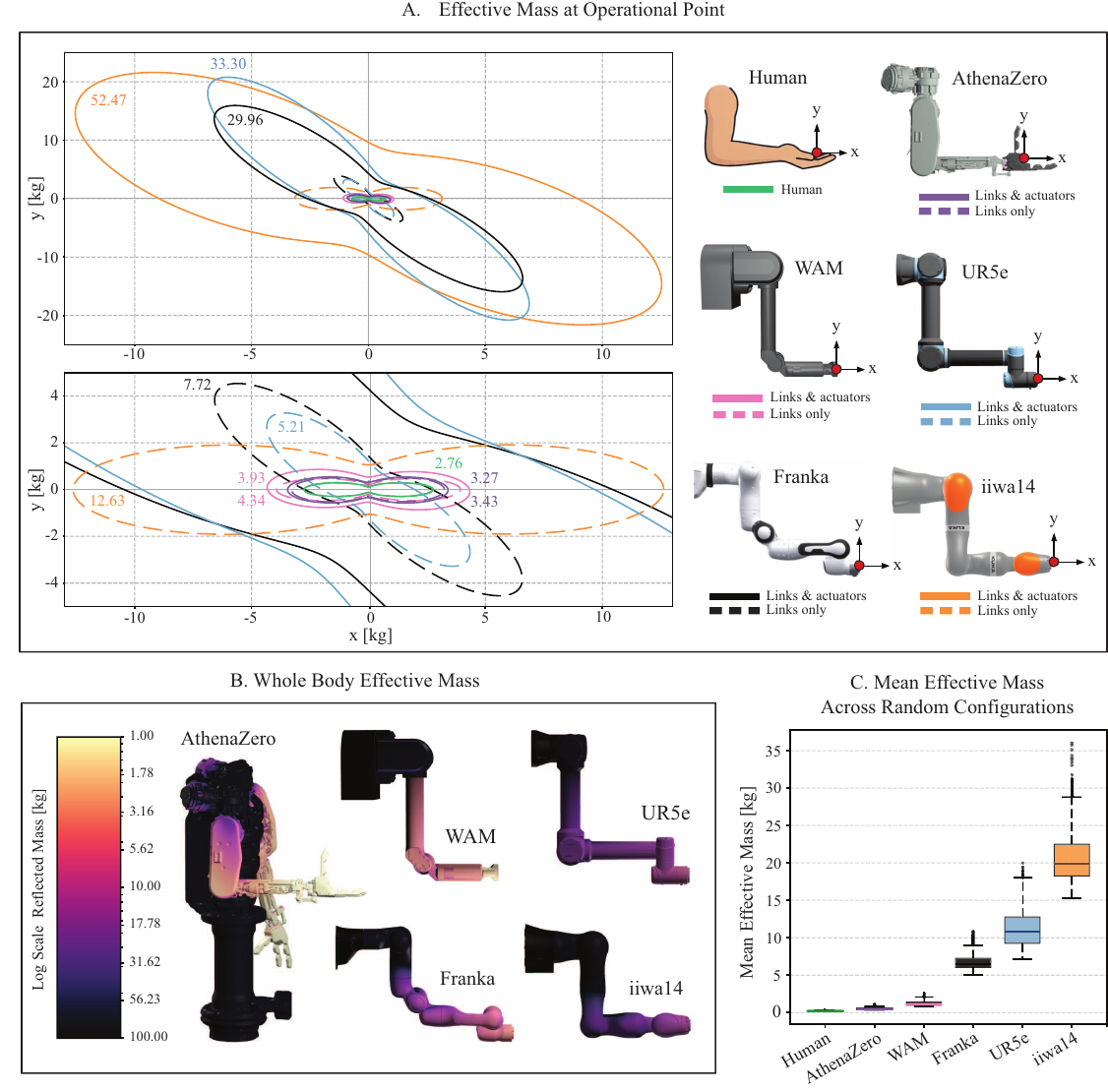}
\caption{\textcolor{editcolor}{\textbf{Effective mass analysis of \athenazero{} and off-the-shelf manipulators.} We compare the effective mass of a human arm, \athenazero{}---with a locked torso---and a set of commercially-available manipulators (Barrett WAM, Franka Research 3, UR5e, and KUKA LBR iiwa 14). \textbf{A.} We compute effective mass in a neutral configuration for an operational point at the wrist across a planar sweep of directions. Solid lines report total effective mass (including reflected inertia of the actuators) whereas dashed lines report the effective mass of the links alone. Numerical values for the maximum effective mass across swept directions for each arm are shown. \textbf{B.} To promote whole-body contact, we compare the effective mass along all body structures, computed in the direction normal to the surface at each point. QDD manipulators (\athenazero{} / WAM) elicit a similar effective mass on their upper arms as the high-geared manipulators do at the end-effector. \textbf{C.} Effective mass is configuration dependent. {\color{editcolor3}We report the distribution of mean effective mass at the wrist across 2000 random configurations. Box limits indicate the 25th and 75th percentiles, the central line represents the median, and whiskers span the range of non-outlier data, extending to the most extreme values within 1.5 times the interquartile range of the box limits. Points falling beyond the whiskers are plotted individually as outliers. }}}

\label{fig:impedance}
\end{figure*}

\begin{figure*}[!p]
\centering
\includegraphics[width=\textwidth,height=0.2\textheight,keepaspectratio]{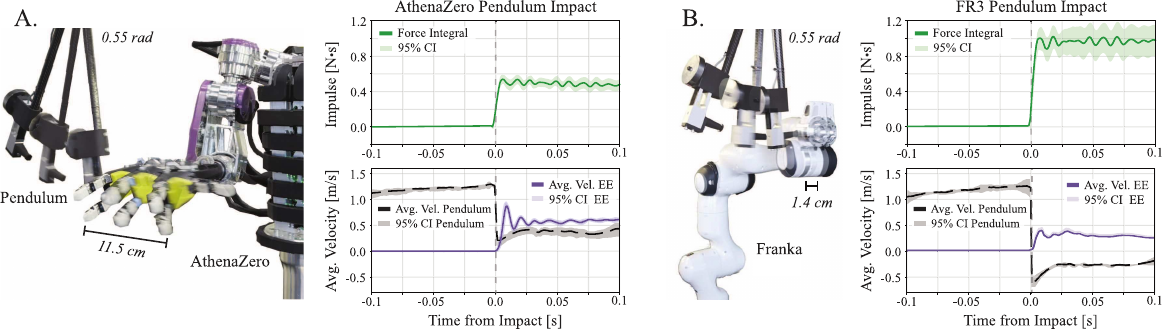}
\caption{\textcolor{editcolor}{\textbf{Pendulum impact test.} \textbf{A.} A 1~kg pendulum impact test on \athenazero{} results in a relatively large end-effector displacement of 115.3~mm and a positive post-contact pendulum velocity. \textbf{B.} A 1~kg pendulum impact test on on the FR3 results in a much smaller end-effector displacement (13.7~mm) and a negative pendulum velocity (bouncing effect). {\color{editcolor3} Error bars indicate 95\% confidence intervals calculated from the standard error of the mean for five independent trials per robot.} }}
\label{fig:impact_test}
\end{figure*}

The results for both \athenazero{} and the WAM were similar to a human arm. By remoting all motors, the WAM achieves a post-shoulder mass (4.81~kg) similar to the human reference (4.48~kg). \athenazero{} similarly prioritizes proximal mass concentration but avoids joint-crossing transmissions (except for the finger motors) to maintain simplicity. Consequently, its post-shoulder mass is higher (5.99~kg), though its link-only effective mass remains low as the lightweight distal links dominate effective end-point mass. The FR3’s post-shoulder mass is even higher (11.06~kg), and its effective mass is substantially greater as its actuators are coincident with the joints. This gap widens when considering reflected rotor inertia, which more than doubles the total effective mass of highly-geared arms. In contrast, \athenazero{} is minimally {\color{editcolor3}affected}, confirming that its inertia is dominated by the links rather than the actuators.

Figure~\ref{fig:impedance}B extends the analysis to operational points along the entire arm for the same configuration. Here, the advantages of QDD become even more evident. {\color{editcolor3}Whereas} low end-point mass can be achieved using series compliance on a geared arm, QDD designs reduce effective mass for contacts at any point on the structure. The heatmaps illustrate that \athenazero{} and the WAM maintain low effective mass even for proximal contacts, with upper-arm values remaining comparable to the distal links of the FR3, UR5e and iiwa14. This makes \athenazero{} well-suited for whole-body manipulation.

Finally, we confirmed the trends held across various configurations. Fig.~\ref{fig:impedance}C presents the distribution of mean effective mass \cite{steinecker2022mean} across 2000 random, non-singular configurations for an operational point at the wrist. Because this metric averages across all directions, values are generally lower than the directional maxima in Fig.~\ref{fig:impedance}A. We noted that \athenazero{} exhibited a slightly lower overall distribution than the WAM. This is due to \athenazero{}’s pitch-yaw wrist configuration, which yields lower effective mass than the WAM’s pitch-roll structure in the direction orthogonal to the plane used in Fig.~\ref{fig:impedance}A.
}

\textcolor{editcolor}{To support our theoretical analysis, we performed pendulum impact tests for both \athenazero{} and the FR3 (Fig.~\ref{fig:impact_test}, {\color{editcolor3}Movie s2}). Both arms, operating in gravity compensation mode, were placed in neutral configurations and impacted from the side of the ``wrist" (the z-direction in Fig.~\ref{fig:impedance}A). Experiencing a measured impulse of  $J = 0.48 \text{ N}\cdot\text{s}$, \athenazero{} yielded to the impact with a resulting velocity of $\Delta v = 0.58 \text{ m/s}$. The measured effective mass was therefore $M_{eff} = 0.83 \text{ kg}$, compared to $M_{eff} = 0.78 \text{ kg}$ computed from the dynamic model.  Conversely, the FR3 experienced an impulse of $J = 0.98 \text{ N}\cdot\text{s}$ and a velocity change of $\Delta v = 0.30 \text{ m/s}$, resulting in an effective mass of $M_{eff} = 3.3 \text{ kg}$ (the dynamic model predicted $M_{eff} = 3.6 \text{ kg}$). This result supports our theoretical analysis as also discussed in \cite{kirschner2021notion}; we attribute the observed discrepancies to pendulum friction, the time resolution of the force sensor, slight uncertainty in the contact location, and unintended contact compliance. Notably, the peak impact force resulting from collision with \athenazero{} ($124.1$~N) was substantially less than that resulting from collision with the FR3 ($208.3$~N).
This tangibly demonstrated the benefit of low-inertia hardware in the event of collision. \athenazero{}’s lower effective mass allowed the pendulum to maintain its forward velocity after collision, whereas collision with the highly-geared FR3 caused the pendulum to reverse direction.
}

\subsubsection*{Stiffness Evaluation}
{\color{editcolor2}
To validate the extent of \athenazero{}'s dynamic range, we conducted a series of experiments validating its active stiffness capabilities. Utilizing the control architecture described in the Supplementary Materials and a calibrated gravity compensation model, we increased the proportional and derivative gains ($\mathbf{k}_p, \mathbf{k}_d$) to a terminal threshold approximately 10\% below the onset of actuator instability. Thereafter, we sent the robot to 30 random configurations in both unloaded and loaded (1.8~kg) payload conditions, yielding the results shown in Fig. \ref{fig:stiffness}.

\begin{figure*}[!htbp]
\centering
\includegraphics[width=\textwidth,height=0.21\textheight,keepaspectratio]{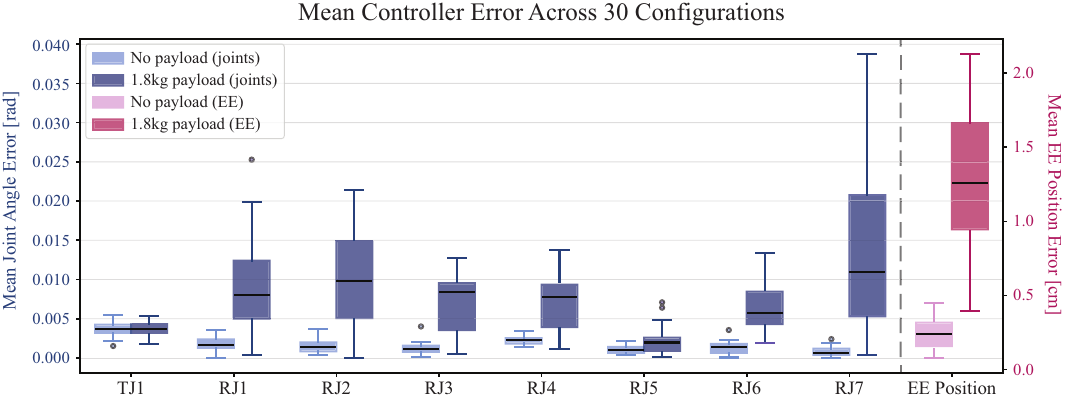}
\caption{\textcolor{editcolor}{\textbf{Robot stiffness evaluation.} Resultant positional offset of \athenazero{} when commanded to 30 random joint configurations in unloaded and loaded conditions. In the loaded condition, the 1.8~kg mass was affixed to the end effector. {\color{editcolor3}Box limits indicate the 25th and 75th percentiles, the central line represents the median, and whiskers span the range of non-outlier data, extending to the most extreme values within 1.5 times the interquartile range of the box limits. Points falling beyond the whiskers are plotted individually as outliers.}}}
\label{fig:stiffness}
\end{figure*}

Our evaluation demonstrated that \athenazero{} exhibited a relatively low kinematic error at the end effector when operating unloaded, averaging approximately~3 mm. Under a load, however, this error increased, resulting in an average deflection of 12~mm across all configurations. An analysis of the joint-level contributions revealed that {\color{editcolor3}Right Arm Joint 7 (RJ7)} was the primary source of this deflection, as it is the lowest torque joint in the chain, yielding an average error of 0.011~rad throughout the trials. As expected, displacement error increased in extended configurations, due to increased lever arms. Overall, these findings indicate that, {\color{editcolor3}although} \athenazero{} maintains adequate stiffness for its intended dynamic tasks, it inherently falls short of the strict millimeter-level precision typical of traditional cobots.

}

\subsection*{Capability Highlights: Throwing, Catching, and Batting}
\label{sec:tasks}

We demonstrated the robot's capabilities on three baseball-inspired tasks: throwing, catching, and batting (Fig.~\ref{fig:catch_throw_bat}), which showcased complex interactions on human-comparable timescales where milliseconds matter (for more information on why these tasks were chosen and the technical approaches taken to achieve them, please refer to the Supplementary Materials). All experiments were conducted in a controlled lab setting, with two \athenazero{} robots vis-à-vis at a distance of 7.3~m. An OptiTrack motion capture suite provided estimated ball positions at 240~Hz for all experiments. Table~\ref{tab:task_success} outlines achieved success for each of these individual skills.

\begin{table*}
    \centering
    \small
    \begin{tabular*}{\linewidth}{@{\extracolsep{\fill}} ccc|c }
        Skill & Maximum Velocity & Accurate Velocity & \\
        \hline
        \hline
        \textit{Throwing}       &  30.8~m/s & 21.4~m/s & \\
         \vspace{0.1cm}
              & Maximum Velocity @ 7.3~m & Reaction Time & Extrapolated Velocity @ 18.4~m \\
        \hline
        \hline
        \textit{Catching}      &  18.3~m/s &  0.398~s & 46.1~m/s \\
        \textit{Batting}       &  13.9~m/s &  0.525~s & 35.0~m/s\\

    \end{tabular*}
    \caption{\textbf{Overview of throwing, catching, and batting capabilities.} We showcase the maximum velocities of each baseball task, in addition to the equivalent catching and batting velocity according to reaction time if the two robots were placed on a standard pitcher's mound and in the batter's box (18.4~m). }
    \label{tab:task_success}
\end{table*}

To highlight, our fastest throw with \athenazero{} was in the form of a tennis ball ($\sim$57~g) at 30.8~m/s in a one-arm configuration. This achievement, to our knowledge, is by far the fastest ball thrown by an anthropomorphically-inspired robot, and equates to approximately 112~kph or 69~mph. A major challenge in throwing accurately at this speed was unexpected slip before desired release. Focusing on release timing, the two-armed system was slowed down, and we found it could throw a baseball ($\sim$145~g) accurately at 21.4~m/s within a 0.25~m x 0.25~m plane at 7.3~m (Fig. \ref{fig:catch_throw_bat}).

\begin{figure*}[!htbp]
\centering
\includegraphics[height=4cm]{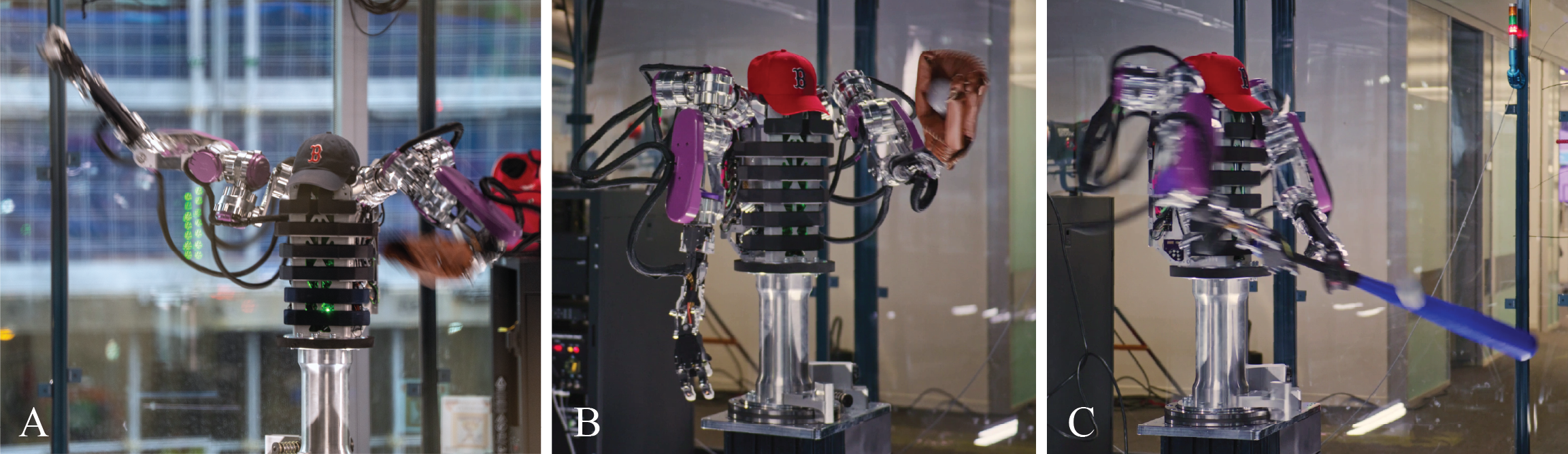}
\caption{\textbf{Baseball-inspired tasks to showcase dynamic manipulation.} \textbf{A.} Throwing a baseball at 13.4~m/s. \textbf{B.} Catching a baseball at 13.4~m/s @ 7.3~m (0.544~s reaction time). \textbf{C.} Batting a tennis ball at 13.4~m/s. @ 7.3~m  (0.544~s reaction time).}
\label{fig:catch_throw_bat}
\end{figure*}

Catching at high speeds was difficult given the short 7.3~m distance between the two robots, substantially less than the standard distance between a pitcher's mound to home plate at 18.4~m. Moreover, estimating a ballistic trajectory takes at least 3 sampled positions in Cartesian space, which further shortens the reaction time window---sometimes drastically according to the sample rate. Within this controlled lab setting, catching of a baseball was successful at speeds up to 18.3~m/s (0.398~s reaction time), which equates to an extrapolated catch at 46.1~m/s (164~kph/102~mph) from the pitcher's mound. Similarly, solid bat contact of a tennis ball where the ball was returned to the opposite robot was achieved with maximum throw velocities of 13.9~m/s (0.525~s reaction time), which is equivalent to 35.0~m/s (125~kph/78~mph) in reaction time from a standard pitcher's mound.

\subsection*{Playing a Game of Catch}
\label{sec:game_of_catch}
The system's capability of generating fast, dynamic, contact-rich behavior was validated through a benchmark task requiring two robots to play a game of catch with a baseball at a human-level cadence. This served as a rigorous testbed for robustness and real-time adaptability.

To start, the robot had to track the baseball to predict its ballistic trajectory, estimating its arrival time at a target plane with an accuracy of about 10~ms. To intercept it, the arm had to complete its motion within a time window of $\leq$0.4~s, seamlessly transitioning to an operational space impedance controller to compliantly absorb the ball's impact. Following a successful catch, the system demonstrated dexterity: the opposite hand had to locate the ball within the glove---adapting to its variable position after each catch---and execute a secure transfer. Proper alignment of the ball in the throwing hand was paramount, as even minor misorientation would propagate into throwing errors, jeopardizing the reciprocal nature of the game. This entire sequence showcased the system's capacity for performing complex, dynamic tasks in an imperfect environment, particularly when a human was introduced into the mix.

\begin{figure}[!htbp]
\centering
\includegraphics[width=\columnwidth,height=0.5\textheight,keepaspectratio]{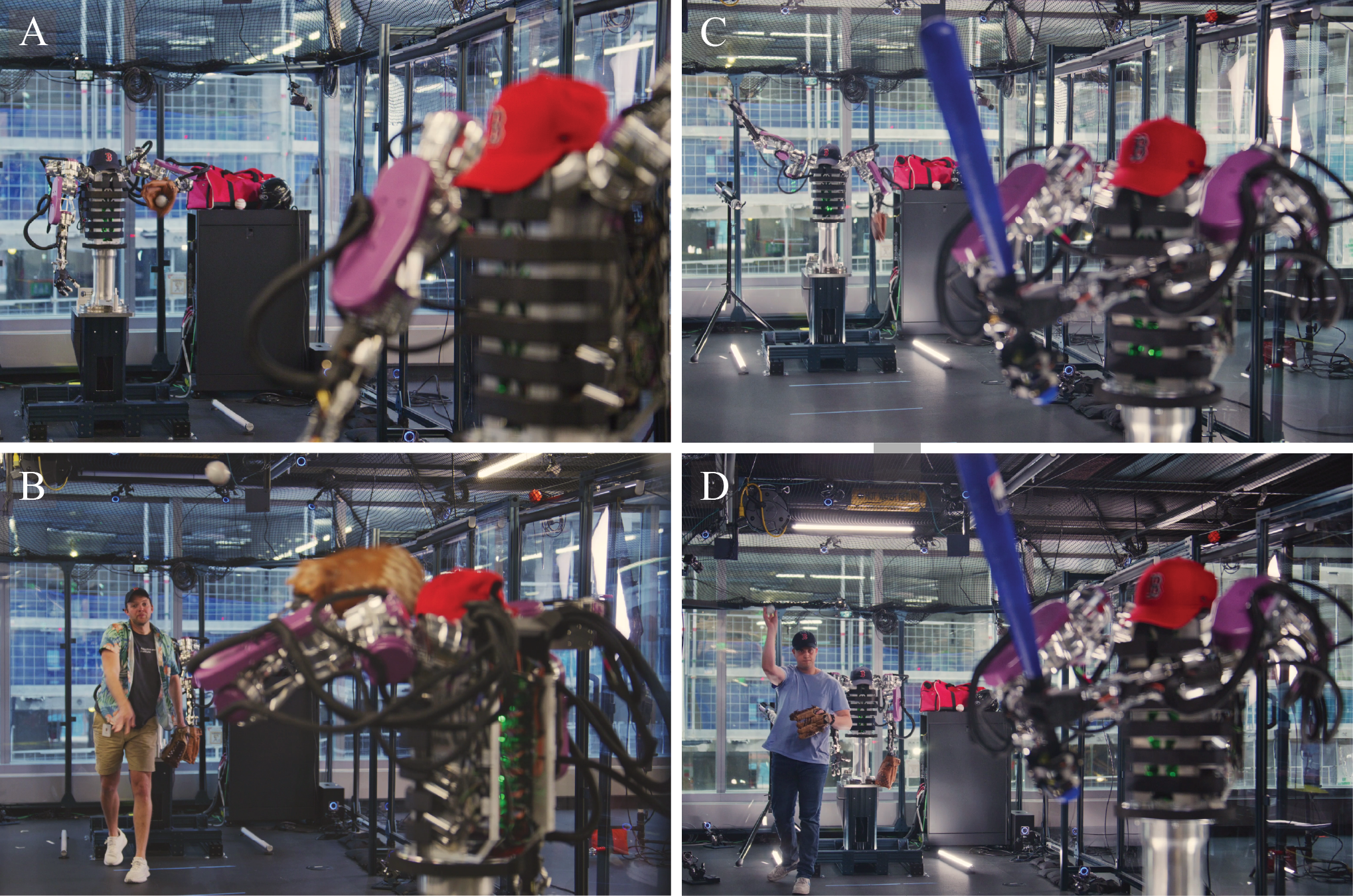}
\caption{\textbf{Cooperative variations of playing a game of catch and participating in batting practice.} \textbf{A.} Two robots playing a game of catch with one another. \textbf{B.} A robot and a human playing a game of catch. \textbf{C.} One robot throwing batting practice to another robot. \textbf{D.} A human throwing batting practice to a robot.}
\label{fig:throw_catch_montage}
\end{figure}

We validated the performance of this task through robot-to-robot and robot-to-human variations (Fig.~\ref{fig:throw_catch_montage}, {\color{editcolor3}Movie s3}). In the robot-to-robot demonstration, throwing speeds were calibrated to approximately 13.4~m/s (0.544~s response time) and total catch-to-throw timings were approximately 11~s. The throwing robot was responsible for throwing a ball to the left side of the catching robot within its reachable workspace. In our evaluation, the two robots were able to volley successfully back and forth up to 8 times. Failure cases were often due to misalignment of the ball transfer after the catch. This in turn correlated to an inaccurate return throw to the opposing robot. To further underscore the adaptability of our system, we replaced one robot with a human, which was successful up to a total of 12 throws back-and-forth.

\subsection*{Batting Practice}
\label{sec:batting_practice}
A subsequent evaluation focused on a different dynamic adaptation task, robot batting practice ({\color{editcolor3}Movie s4}). In this scenario, a tennis ball was thrown, tracked, and batted by the robot sequentially---that is, the bat was not dropped, or regrasped, during the entirety of the batting process. This task was especially challenging because it required both the robot's intrinsic mechanical accuracy and millisecond-level timing to make successful contact. Specifically, the temporal window for making effective contact was narrow; executing the swing just 2~ms too early or too late would result in a miss. Consequently, the system could not rely on a pre-programmed motion. Instead, it continuously adapted its swing in real time based on an estimated projection of the ball within a 2D ``strike zone" (0.4~m x 0.7~m on the robot's left side). Only through a tight integration of these mechanical and computational capabilities could the robot successfully strike the ball at appreciable speeds.

This benchmark was demonstrated in two configurations, both robot-to-robot and human-to-robot (Fig. \ref{fig:throw_catch_montage}). In the robot-to-robot variation, the throwing robot attempted to deliver a pitch randomly within the ``strike zone" at approximately 13.4~m/s (0.544~s response time). If the location of a pitch was outside the ``strike zone", the batting robot did not swing. Subsequent throws were made from the pitching robot approximately every 12~s. In our evaluation, the two robots were able to participate in batting practice back and forth for approximately 3 minutes, with a combination of ``foul balls" and ``grounders" hit back to the pitcher. Similar to playing a game of catch, we also replaced one robot with a human that attempted to throw the ball within some variation of 13.4~m/s $\pm$ 2~m/s. Here, we found the robot to successfully connect with the pitch with similar success rates to the robot-to-robot variation.


\section*{Discussion}
\label{sec:discussion}

\subsection*{Limitations}
{\color{editcolor}
The \athenazero{} project evaluates to what extent applying contemporary motor and compute technology can alleviate historically-known limitations of quasi-direct drive actuators. {\color{editcolor3}Although} the results are promising, there are remaining challenges that require further investigation. Foremost is the susceptibility to thermal saturation (overheating) under sustained static loads. To expedite research into core manipulation challenges, active thermal mitigation was deprioritized in favor of software safeguards, which are configured to trigger when motor temperatures exceed 80° C. A thermal model was developed to predict thermal excitations, which was leveraged in offline simulation. Operationally, this necessitates control strategies distinct from those of traditional arms. For instance, task execution cannot rely on the prolonged static holding of payloads exceeding 2 kg at full extension. {\color{editcolor2}Furthermore, the transparency and backdrivability characteristics that enable impedance matching also reduce endpoint stiffness, rendering conventional trajectory-tracking difficult. Instead, these hardware constraints closely mirror human physical traits, naturally motivating a shift toward more adaptive, human-inspired interaction strategies.
}

A number of additional deliberate engineering tradeoffs were accepted to enable rapid development. The first was the use of off-the-shelf motor drivers, which restricted actuator currents to 70~A, despite the motors' 130~A peak rating at 48~V. This constraint capped the available torque, inherently limiting the robot's dynamic performance during impulsive maneuvers, such as throwing. Industrial design refinements were also considered secondary. Consequently, external cable routing and reflective surface finishes occasionally interfered with optical motion capture systems. Finally, in our most radical attempt to minimize distal mass, we actuated the hands via long Bowden cables with actuators housed in the torso. {\color{editcolor3}Although} this successfully reduced end-effector inertia, the resulting transmission friction restricted the maximum grasping force and the dynamic capabilities of the hands. Overall, these design compromises were deliberately chosen to prioritize rapid hardware iteration, enabling the team to isolate and evaluate the core architectural insights presented in this article.
}

\subsection*{System Highlights}

Our analysis demonstrates that \athenazero{} achieves an effective mass approximately an order of magnitude lower than conventional collaborative robots, bringing it substantially closer to human biomechanical baselines. As anticipated, its effective mass is comparable to other low-inertia manipulators, such as the WAM and AMBIDEX. However, \athenazero{} attains this performance without relying on the complex tendon-routing inherent to those systems. Although effective mass is intrinsically configuration-dependent, this fundamental reduction is maintained throughout the workspace, driven by the inertial scaling laws of the motor gearing.

To evaluate dynamic performance, we executed three baseball-inspired tasks using both the single- and dual-arm configurations of \athenazero{}. The single-arm system achieved release velocities of approximately 30.8~m/s with a tennis ball ($\sim$57~g) and 21.4~m/s with a standard baseball ($\sim$145~g). For the dual-arm configuration, which required accurate targeting (within a 25~cm × 25~cm bounding box) at distances $\ge$~5~m, the system reliably threw a baseball at 21.4~m/s. From an algorithmic perspective, we observed that the direct collocation techniques used for trajectory optimization were highly sensitive to cost and constraint shaping; iteratively introducing objectives into the solver was critical for convergence. Because these trajectories required $\sim$5 minutes to compute, we subsequently developed an online adjustment tool based on nonlinear kinematic trajectory optimization.

The catching framework demonstrated reliable performance for projectiles thrown from 7.3\,m at velocities up to 13.4\,m/s (30\,mph), provided the predicted intercept remained within the robot’s reachable workspace.  {\color{editcolor3}Although} we recorded successful catches at speeds reaching 18.3\,m/s, system reliability diminished considerably at these upper bounds. This failure mode was primarily driven by the ball rebounding from the palm faster than the fingers could close. We identified this bottleneck as a limitation in the stability of our closed-loop impedance control damping bandwidth. Thus, whereas a pure velocity-matching strategy remains unsuitable for high-speed catching, a purely impedance-matching approach also has limitations.

Finally, the batting framework achieved a success rate of over 82\% (27 hits in 33 swings), reliably intercepting balls launched at approximately 13.4~m/s from a distance of 7.3~m to random positions within the predefined ``strike zone". Swinging maneuvers were executed sequentially without requiring the robot to reposition or regrasp the bat between attempts. However, our ability to replicate highly anthropomorphic batting motions was fundamentally constrained by the manipulator's kinematics, specifically the limited range of motion in wrist deviation.

\subsection*{Implications for Manipulation}

\textcolor{editcolor}{
{\color{editcolor3}Although} tasks such as throwing, catching, and batting may initially appear to be niche applications, their underlying mechanics force a fundamental reevaluation of robotic manipulation. Executing these highly dynamic behaviors exposed critical limitations in current motion solvers. Unlike traditional high-inertia manipulators, where planning is often dominated by task kinematics, the inherent dynamics of our system necessitate a shift toward tightly-integrated torque control and gain scheduling techniques. Because QDD actuators exhibit high backdrivability and control authority, minor offsets in desired trajectories (particularly in acceleration) considerably affect feedforward dynamics. Consequently, prioritizing smooth, energy-efficient motion becomes essential. {\color{editcolor2} Crucially, platforms like \athenazero{} provide a physical testbed to explore whether ``natural" motion can be reduced merely to minimizing jerk, or if there are other objectives humans and other animals satisfy.} This indeed aligns with the results from multiple studies since the 1980s, which show that minimum jerk does not explain human motion in general \cite{MotorMemory2010,Engelbrecht2001}.}

\textcolor{editcolor}{
Ultimately, generalized manipulation remains a formidable system-level challenge; unifying control, sensing, and cognition into a single framework is an open frontier. This work represents our initial foray into the control and sensing pillars of that triad. Because real-world environments are inherently unpredictable, robotic systems must gracefully manage the indescribables: the inevitable ambiguities in contact friction, object mass, shifting geometry, and transient dynamics, for example. We believe that approaching core challenges from a dynamics perspective will allow researchers to use the right tools when asking the right questions about manipulation.
}
\section*{Materials and Methods}
\subsection*{Robot Hardware}
\subsubsection*{Actuators}
Actuator design reigns as a central ingredient to execute our design methodology. To minimize reflected inertia and maximize transparency, we utilize quasi-direct drive motors. \textcolor{editcolor}{ Since we aim for human-like mass properties, we concurrently require human-like torques, deriving our targets from the isometric torque production of equivalent human joints \cite{mayer1994normal, kotte2018normative, yoshii2015measurement}. Based on these benchmarks, pointing towards approximately 90~Nm for the shoulder joints, we selected frameless stator-rotor pairs using specific peak torque (Nm/kg) \cite{seok2012_cheetah1} as our primary performance metric, bounded by actuator space constraints and the availability of single-stage, low-ratio gearsets ($\leq$10:1).} Notably, we excluded thermal performance metrics from our selection criteria, as our intended use cases involve highly dynamic, high-acceleration motions at low duty cycles \cite{kenneally2016_minitaur}. We similarly disregarded metrics incorporating the mechanical time constant, as we expect the total system inertia to be dominated by the structural links rather than the motors themselves \cite{urs2022_metrics}.

For the torso, shoulder, and upper arm, we chose the Allied Motion Megaflux motors \cite{allied_motion}. This series of frameless pairs is designed for large peak torques, featuring considerable amounts of backing steel on both the rotor and stator to limit magnetic saturation. We achieve our gear reduction through a single-stage helical planetary gearset ($\leq$8e-4~rad backlash), as the load-sharing capabilities of planetary reductions and the increased strength of helical gears minimize both gearbox volume and mass. A full-complement crossed-roller bearing supports the output of each motor, where the rotor is supported between bearings in the carrier and the rear housing plate. Due to similar torque requirements, we use a 95~mm diameter actuator to drive all three shoulder joints (J1, J2, J3) and the elbow flexion-extension joint (J4). Two actuators are used in parallel to drive the torso via an additional belt reduction. The elbow pronation-supination joint (J5) uses a slightly smaller 76~mm diameter actuator built under the same QDD philosophy. Motor selection and transmission characteristics for the wrist and fingers are described in the sections on wrist and hand design, respectively. Table \ref{tab:joint_specs} presents joint specifications for the entire robot.

{\color{editcolor}
All motors are governed by Copley Nano NES motor drivers utilizing nested position, velocity, and current control loops in CSP mode \cite{copley_canopen_2018} (see Supplementary Materials for more information). To reduce distal mass, the physical drivers are housed in the robot's torso, with cable bundles routed outward to each motor. The position and velocity control loops operate at 4~kHz, whereas the current control loop runs at 16~kHz. Over a single EtherCAT bus operating at 1~kHz, all motor drivers receive commands for position, setpoint velocity, feedforward torque, position gains, and velocity gains. Ultimately, this capacity for online, high-frequency gain scheduling enables precise control over the motor's output impedance.
}

\subsubsection*{Shoulder and Elbow} \label{sec:arm}
The shoulder utilizes three structural 95~mm diameter motors with a 5:1 gear reduction, arranged in a standard roll-pitch-roll configuration (Fig.~\ref{fig:shoulder_wrist_hand}). The axes for all three motors coincide at a single point, nominally acting as a spherical joint. However, this shoulder reaches a singularity when the axes of the first and third motors become collinear. To mitigate the affect of this singularity on the system's overall dexterity, we offset the first shoulder joint by 20~degrees upward and 30~degrees forward. This kinematic offset substantially improved the feasibility of the throwing motion, which we validated by tracking human pitching trajectories recorded via motion capture.

The upper arm and elbow present the first practical opportunity to utilize a transmission to reduce link inertia by shifting an actuator proximally. The elbow consists of two orthogonal joints: flexion-extension (J4) and pronation-supination (J5). After evaluating various methods for coupling or relocating both actuators, we struck a balance between link inertia and mechanical simplicity. We relocated the 95~mm diameter J4 motor proximally using a timing belt transmission, with leaving the 76~mm diameter J5 motor collocated directly at the joint. Both motors utilize a 5:1 reduction.

\subsubsection*{Wrist}
\label{sec:wrist}
The \athenazero{} wrist is a two DoF parallel mechanism driven by prismatic actuators (Fig. \ref{fig:shoulder_wrist_hand}). Here, the penalty for distal mass is larger and requires the use of a more complicated remoting mechanism.
The wrist is actuated by two 38~mm diameter frameless pairs from TQ \cite{tq} driving 8~mm pitch ballscrews. At the wrist's nominal position, each motor has a reduction ratio of 10.6:1 in flexion and 7.5:1 in deviation, resulting a total peak torque of $\sim$16.1~Nm in flexion and $\sim$11.4~Nm in deviation.

The wrist mechanism is a 2-SS[RRR\underline{P}] + 1RR parallel mechanism, which is most similar to the RH5 Manus wrist (2-SU[RR\underline{P}R]+1U) \cite{boukheddimi2022_rh5manus}. Compared to the RH5 Manus wrist, the prismatic actuator is grounded, and wire management is simplified. A simple and thus high strength leg in the parallel mechanism withstands the bulk of the forces on the hand. This strength addresses concerns with mechanisms like the Quaternion wrist \cite{kim2018_quatwrist} and the DLR wrist \cite{yoon2021_dlrwrist} upon impact. A central RR link, which can be implemented with a hollow central cavity, makes it easy to route Bowdens for the hand tendons and wires for communications close to the center of rotation of the wrist. This mechanism has a total of 4 DoFs: flexion, deviation, and two uncontrolled rotations of the ``coupler" links that connect the spherical joints. These uncontrolled rotations can be removed by replacing one of the spherical joints with a universal joint. However, these DoFs do not affect the flexion or deviation degrees of freedom and enable the use of simple spherical bearings.

The parallel architecture of the wrist results in forward kinematics that lack a simple closed-form analytical solution. However, the inverse kinematics problem is analytically solvable through sphere-circle and line-circle intersections \cite{tsai1999robot}, commonly referred to as trilateration. To circumvent the computational cost of numerical solvers for the forward problem, we utilize the analytical inverse solution to generate high-density lookup tables for both forward/inverse kinematics and Jacobians. By employing bilinear interpolation, we achieve $O(1)$ computational complexity with query times of approximately 3~$\mu$s. This microsecond-scale performance ensures that conversions remain negligible within our 1~kHz control loop, while maintaining interpolation residuals below the mechanical resolution of the system.

\subsubsection*{Hand}
\label{sec:hand}
The \athenazero{} end-effector is a three-finger, Dollar-style \cite{dollar2010highly} tendon-driven underactuated hand (Fig. \ref{fig:shoulder_wrist_hand}). The three underactuated fingers interdigitate, with link lengths optimized for power grasping both a baseball ($\sim$7.4~cm diameter) and a baseball bat ($\sim$2~cm at the handle) upon closure. The relative proximal-to-distal stiffnesses of each finger joint follows an approximate 1:1.4 ratio for desired free-swing closing patterns \cite{bircher2019design}. Each finger is driven by a single 25~mm diameter frameless pair from TQ \cite{tq} coupled to an 18.8:1 gear reduction. Similar to other motors on the robot, position, velocity, feedforward torque, position gain, and velocity gain setpoints are sent to the motor drivers as 1~kHz. The actuators are remoted into the torso of the robot to minimize mass at the end of the arm. Force is transmitted to the fingers through a 1.8~m long Bowden transmission shielded by aluminum links to resist buckling, using 0.8~mm diameter Dyneema tendons. This simple hand design was focused on minimizing end-of-arm mass while maintaining appreciable torque ($\sim$4~Nm peak) and speed ($\sim$80~ms closing time) of the fingers.

The fingerpads and palm pad are constructed of TEPU 30A~\cite{inkbit}, which house barometer-based tactile sensors that operate at 200~Hz. The embedded barometers measure pressure from local contact. Eight sensors are housed in the palm and three in each of the fingertips, providing a nominal 3~mm resolution on contact location and approximately 0.1~N resolution on force. Local compute for these sensors are housed in the back of the hand. Examples of tactile sensing utility is provided in Movie s5.

\subsection*{Inertial Properties}
\subsubsection*{Theoretical Analysis of Effective Mass}
Effective mass serves as an intuitive proxy for evaluating the inertial characteristics---or the resistance to change in motion---of a system. This value represents the apparent mass of a manipulator as reflected through its kinematics, i.e, this value is configuration dependent. When a wrench is exerted at an operational point on the robot, the effective mass relates that action with the resultant acceleration. Intuitively, units of effective mass are in $kg$, instead of standard inertial units of $kg\cdot m^2$.

More formally, we calculate the effective mass of different manipulators as described in \cite{khatib1995inertial}. Let's assume we have a fully-actuated robot with joint configuration $\textbf{q}$. Given an operational point on the robot, we can compute a linear Jacobian, $\mathbf{J_v(q)}$, associated with the linear velocity at that operation point. A representative mass matrix, $\mathbf{M(q)}$, for the robot is a superposition of the link inertia and the actuator inertia values:

\begin{equation}
\mathbf{M(q) = M_l(q) + M_a}
\end{equation}

\noindent Here, $\mathbf{M_l(q)}$ encodes link inertia information and $\mathbf{M_a}$ encodes actuator reflected inertia, which is not configuration dependent. It is straightforward to calculate $\mathbf{M_l}$ through a robot model and traditional techniques. Obtaining $\mathbf{M_a}$, on the other hand, is often overlooked. Here, $\mathbf{M_a}$ is an $n\times n$ diagonal matrix representing the rotor inertia of each actuator. We can write it as:

\begin{equation}
\mathbf{M}_a =
\begin{bmatrix}
I_{r,1} N_1^2 & 0 & \cdots & 0 \\
0 & I_{r,2} N_2^2 & \cdots & 0 \\
\vdots & \vdots & \ddots & \vdots \\
0 & 0 & \cdots & I_{r,n} N_n^2
\end{bmatrix}
\end{equation}
where $I_{r,i}$ and $N_i$ are the rotor inertia and gear reduction for the actuator driving the $i^{th}$ joint, respectively. That the gear ratio contributes through a squared term is a key causality for the high total effective mass of traditional highly-geared manipulators.

Given the translational Jacobian $\mathbf{J_v(q)}$, and the mass matrix of the robot $\textbf{M(q)}$, we can compute the pseudo kinetic energy matrix, $\boldsymbol{\Lambda}_v^{-1}(\mathbf{q})$:
\begin{equation}
    \label{eq:kinetic energy matrix}
    \boldsymbol{\Lambda}_v^{-1}(\mathbf{q}) =
    \mathbf{J}_v(\mathbf{q})\,\mathbf{M}^{-1}(\mathbf{q})\,\mathbf{J}_v^{\mathsf{T}}(\mathbf{q})
\end{equation}

This equation relates the translation of the end-effector to the force felt at the operational point. Given, $\boldsymbol{\Lambda}_v^{-1}(\mathbf{q})$ and a unit vector $\textbf{u}$ describing the direction of an instantaneous perturbation, we can calculate the effective mass, $m_{\textbf{u}}$, which is a scalar value, as:

\begin{equation}
    \frac{1}{{m_{\textbf{u}}(\boldsymbol{\Lambda}_v)}} = \textbf{u}^T\boldsymbol{\Lambda}_v^{-1}(\textbf{q})\textbf{u}
\end{equation}

The plots presented in Fig.~\ref{fig:impedance}A were calculated by sweeping over the parameter $\textbf{u}$ in a 2-dimensional plane. Effective mass depends both on choice of operational point and configuration. To provide an intuitive visualization, we select a point at the wrist---finger joints are not included---and place each robot in a similar elbow-bent configuration. \textcolor{editcolor}{Figure~\ref{fig:impedance}B computes $m_{\textbf{u}}$ for the same configuration, but for operational points along the arm in the direction of the surface normal. To capture configuration-dependence, Fig.~\ref{fig:impedance}C computes the mean effective mass over a sweep of spatial directions and across a large set of sampled joint angles.}

\textcolor{editcolor}{Our analyses of the WAM, FR3, UR5e, and KUKA LBR iiwa14 are based on community-generated robot models, supplemented by manufacturer-provided information when available. The models---containing kinematic and link inertia information, but not rotor inertia and gear reduction values---for the FR3, UR5e, and iiwa14 are sourced from~\cite{menagerie2022github} whereas the WAM model resides in ~\cite{Jianxiang_Mujoco_Mobile_WAM_2022}. These models were used to compute $\mathbf{M_l(q)}$. Reflected actuator inertia for the FR3 and iiwa14 are from~\cite{robotlocomotion_models}. Rotor inertia and gear ratio values for the UR5e are provided in \cite{porcelli2020dynamic}, and validated directly from a Universal Robots control box. Barrett provides rotor inertia and gear reduction values in their publicly available documentation \cite{barrett2008wam}. The joint configurations used for the analysis can be found in Table \ref{tab:ef_mass_joints}. To ensure these results are easily reproducible, we have compiled the dynamic models and actuator inertia information for all robots analyzed in an open-sourced repository~\cite{raiinst2026effectivemass}. }

\begin{table*}
    \centering
    \begin{tabular*}{\linewidth}{@{\extracolsep{\fill}} cccccc }
        Joint Angle [rad] & \athenazero{} & WAM & FR3 & UR5e & iiwa14 \\
        \hline
        \hline
        \textit{$q_1$} & -0.1396 & 0 & 0 & 0 & 0\\
        \textit{$q_2$} & 0.1396 & -1.5708 & -1.5708 & -3.14159 & -1.5708\\
        \textit{$q_3$} & 0.5236 & 0 & 0 & 1.5708 & 0\\
        \textit{$q_4$} & 1.5708 & 1.5708 & -1.5708 & 3.14159 & -1.5708\\
        \textit{$q_5$} & 1.5708 & 0 & 0 & -1.5708 & 0\\
        \textit{$q_6$} & 0 & 0 & 3.14159 & 0 & 0\\
        \textit{$q_7$} & 0 & 0 & 0.5236 & - & 0 \\
    \end{tabular*}
    \caption{\textbf{Neutral robot joint configurations.} The specified joint angles place compared robots in similar poses for effective mass analysis.}
    \label{tab:ef_mass_joints}
\end{table*}

\subsubsection*{Empirical Analysis of Impact Response}
\label{sec:impulse_def}
{\color{editcolor}
Empirical analysis validated our theoretical model and connected computations to tangible collision outcomes (Movie s2). To achieve this, we developed an instrumented carbon fiber pendulum featuring a 1~kg mass, a 0.5~m shaft, a calibrated force sensor, and an affixed 16-bit encoder. The FR3 and \athenazero{} were placed in a nominal rest configuration under gravity compensation mode. Data from the encoder, force sensor, and robots were queried at 1 kHz. The pendulum was raised to approximately 0.55~rad before release, and five cycles were completed per robot. During these experiments, we collected the robots' joint configurations and velocities, alongside the pendulum's position, velocity, and force data.

Across the five trials, the data demonstrated high repeatability. Using the data described in the Results section, we calculate the effective mass, $M_{eff}$, using:

\begin{equation}
    M_{eff} = \frac{\int F dt}{\Delta v} = \frac{J}{\Delta v}
\end{equation}
where $J$ is the force integral of the pendulum over time and $v$ is the resultant release velocity of the operation point.

For the sake of completeness---though independent of the effective mass analysis---a strict accounting of the total system momentum must acknowledge that a considerable amount of momentum is transferred to the ground via the physical constraints (supports) of the experimental setup. However, the contact impulses ($0.48 \text{ N}\cdot\text{s}$ and $0.98 \text{ N}\cdot\text{s}$) and the pre-impact relative velocity ($v_{rel} = 1.2 \text{ m/s}$) are Galilean invariants across all inertial frames. Therefore, while explicitly tracking the total momentum dissipation within the laboratory frame provides a full systemic overview, it is not required for, nor does it alter, the fundamental effective mass analysis.
}

\subsubsection*{Statistical Analysis}
{\color{editcolor3}
Quantitative results were summarized through descriptive aggregation of model-based computations and repeated physical trials. Effective mass and stiffness performance were characterized over randomly sampled joint configurations (2000 non-singular configurations per manipulator in Fig.~\ref{fig:impedance}C; 30 configurations per loading condition in Fig.~\ref{fig:stiffness}), and summary values quoted in the text are arithmetic means over those samples. Pendulum impact quantities, including impulse, velocity change, and effective mass, were computed per trial and summarized as the arithmetic mean over the trials completed per robot. Task success rates were computed as the empirical proportion of attempts meeting the task-specific success criterion, whereas velocities, reaction times, and volley counts denote maximum observed values.

Box limits indicate the 25th and 75th percentiles, the central line the median, and whiskers the range of non-outlier data; error bars in Fig.~\ref{fig:impact_test} denote 95\% confidence intervals from the standard error of the mean and were used to visualize trial-to-trial repeatability. These descriptive summaries were not used for inferential comparison; no statistical hypothesis tests, $P$~values, or significance criteria were computed. Figure-specific sample sizes and evaluation protocols are provided in the corresponding figure captions and the Supplementary Materials. Data aggregation, descriptive-statistic computation, and visualization were performed using Python with NumPy and Matplotlib.
}

\newpage


\clearpage 

\bibliography{bib}

\section*{Acknowledgments}
\label{sec:acknowledgements}
The authors would like to thank Marc Raibert, Jessica Hodgins, and Christopher Atkeson for their insightful guidance throughout this project.
\paragraph*{Funding:} This research was supported by the AMP/CONDOR/APEX teams at the RAI Institute, with funding from the Hyundai Motor Group.
\paragraph*{Author Contributions:}
A.S.M. led technical analysis and control of the system. G.X., H.B., and P.T. led design, analysis, and fabrication of the custom actuators and mechanisms. A.S.M., M.Y., and M.A.E. contributed the framework related to throwing. C.B. and R.T. developed the impedance-matching and catching framework. C.W. and E.Sh. spearheaded the realization of closed-chain control for batting. E.Sh. also led the design of the interaction-control framework. O.A., O.D.Y., J.O.M, and A.W. developed the on-board tactile sensors. J.A. contributed to finger design for throwing and catching. M.B. and M.W. built the ball tracking framework used for estimating ballistic trajectories. V.D.D., S.K., and E.So. provided electrical support for power distribution, safety, and sensing. F.H. contributed to deterministic control loop modifications. A.Moz. led technical system-level design. A.A.R., A.Moz., N.R., and L.O. provided direction and technical insights throughout the duration of the project.
{\color{editcolor3}\paragraph*{Competing Interests:}
A.S.M. holds a patent in synchronous multi-actuator control. G.X. holds a patent on parallel mechanism design and optimization. C.W. and E.Sh. hold patents on phase-based control. }
{\color{editcolor3}\paragraph*{Data, code, and materials availability:}
All data required to evaluate the conclusion of this paper are available at \url{https://zenodo.org/records/21939225}, in the project repository \url{https://github.com/rai-opensource/effective_mass_analysis}, or in the Supplementary Materials.  No new materials were generated or used in this study. }


{\color{editcolor3}\subsection*{Supplementary materials}
Supplementary Methods\\
Supplementary Figures s1 to s2\\
Supplementary Movies s1 to s5\\
Supplementary Movie Captions\\}







\clearpage
\newpage

\setcounter{table}{0}
\renewcommand{\thetable}{S\arabic{table}}
\setcounter{figure}{0}
\renewcommand{\thefigure}{S\arabic{figure}}
\setcounter{equation}{0}
\renewcommand{\theequation}{S\arabic{equation}}
\setcounter{page}{1}




\clearpage
\onecolumn
\begin{center}
  {\fontsize{18}{22}\selectfont \sffamily \bfseries Supplementary Materials for\par}
  \vspace{0.4em}
  {\fontsize{14}{18}\selectfont \sffamily \bfseries \color{color2} \scititle \par}
  \vspace{1.5em}

  {\sffamily\fontsize{9.5}{13.5}\selectfont
    Andrew~S.~Morgan,$^{1\ast}$
    Gregory~Xie,$^1$
    Capprin~Bass,$^1$
    Rachel~Thomasson,$^1$
    Chunpeng~Wang,$^1$
    Erfan~Shahriari,$^1$
    Harrison~Busa,$^1$
    Oluwaseun~Araromi,$^1$
    Joseph~Aronov,$^1$
    Michael~Burgess,$^1$
    Velin~D.~Dimitrov,$^1$
    Matthew~A.~Estrada,$^1$
    Faris~Hamdi,$^1$
    Samuel~Kendig,$^1$
    Taeyoon~Lee,$^1$
    Jose~Oscar~Mur-Miranda,$^1$
    Emma~Sommers,$^1$
    Paul~Titchener,$^1$
    Margaret~Wang,$^1$
    Achu~Wilson,$^1$
    Mark~Yeatman,$^1$
    Osman~Dogan~Yirmibesoglu,$^1$
    Alfred~A.~Rizzi,$^{1\dagger}$
    Annan~Mozeika,$^{1\dagger}$
    Nicolas~Rojas,$^{1\dagger}$
    Lael~Odhner$^{1\dagger}$
  \par}
  \vspace{1em}

  {\sffamily\fontsize{8.5}{11}\selectfont \color{gray!80!black}
    $^1$RAI Institute, Cambridge, Massachusetts, 02142, USA\\[0.4em]
    \normalcolor
    $^\ast$Corresponding author: \texttt{andy@rai-inst.com} \quad\textbar\quad
    $^\dagger$Co-Principal Investigators
  \par}

  \vspace{1.2em}
  \hrule height 0.6pt
\end{center}
\vspace{1.5em}

\subsubsection*{This PDF file includes:}
Supplementary Methods \\
Supplementary Movie Captions

\subsubsection*{Other Supplementary Materials for this manuscript:}
Figures s1 and s2 \\
Movies s1 to s5
\newpage

\subsection*{Materials and Methods}
\label{sec:SP_materialsmethods}

\subsubsection*{Control Architecture}
\label{sec:control}

\begin{figure}
\centering
\includegraphics[width=1\textwidth]{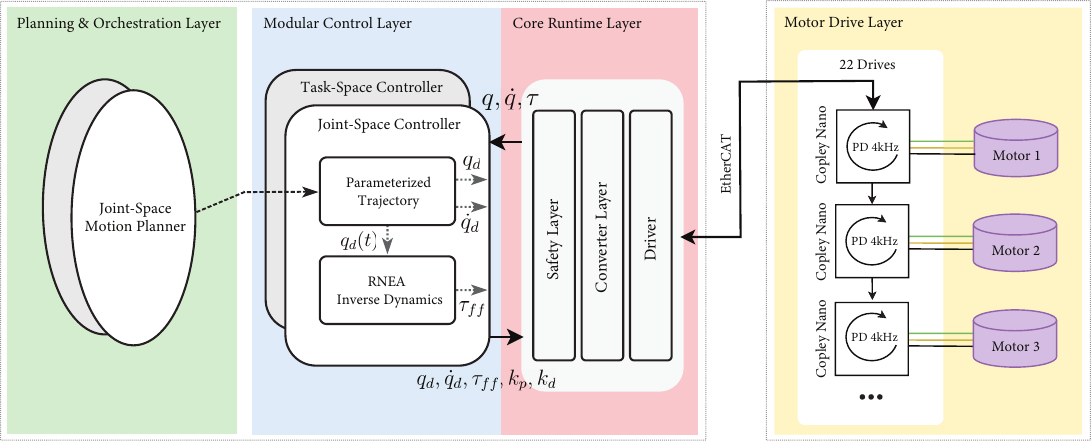}
\caption{\textcolor{editcolor}{\textbf{Layered control architecture of \athenazero{}.} High-level planners generate motion and control parameters that are evaluated within a deterministic 1~kHz real-time loop to produce joint- or task-space reference signals. These references propagate through the control and runtime layers and are ultimately translated into high-frequency actuator commands over EtherCAT.}}
\label{fig:control blocks}
\end{figure}
{\color{editcolor}
The control architecture of \athenazero{} is designed to maximize transparency and bandwidth by distributing the control effort between a real-time host and high-frequency control loops on the motor drives. To make the design easier to interpret, we distinguish four layers in the control architecture (see Fig. \ref{fig:control blocks}), described individually in the following: Motor Drive, Core Runtime, Modular Control, and Planning \& Orchestration.

\textbf{Motor Drive Layer:} Central to the \athenazero{} control architecture are the distributed Copley Nano motor drives communicating with the host computer over an EtherCAT bus. Deterministic EtherCAT communication enables synchronous command and feedback exchange; in our system, the round-trip communication latency was measured to be approximately 650~$\mu$s, allowing the host control loop to operate reliably at 1~kHz.

The motor drives operate in Cyclic Synchronous Position (CSP) mode \cite{copley_canopen_2018}. Although internally implemented using nested position, velocity, and current loops, the actuator-level behavior can be interpreted as a joint-space proportional–derivative controller with feedforward torque. Command references consisting of desired joint position and velocity $\mathbf{q}_d, \dot{\mathbf{q}}_d  \in \mathbb{R}^n$, feedforward torque $\boldsymbol{\tau}_{ff}~\in~\mathbb{R}^n$, and gain vectors $\mathbf{k}_p, \mathbf{k}_d \in \mathbb{R}^n_{\geq 0}$ are transmitted from the host computer to the drives via EtherCAT, whereas measured joint position velocity $\mathbf{q}, \dot{\mathbf{q}}  \in \mathbb{R}^n$ are returned through the same communication cycle. The resulting actuator torque command can therefore be expressed as:

\begin{equation}
    \boldsymbol{\tau}_{cmd} = \boldsymbol{\tau}_{ff} + \text{diag}(\mathbf{k}_p)(\mathbf{q}_d - \mathbf{q}) + \text{diag}(\mathbf{k}_d) (\dot{\mathbf{q}}_d - \dot{\mathbf{q}}) \label{eq:copley}
\end{equation}
By setting $\mathbf{k}_p = \mathbf{0}$ and $\mathbf{k}_d = \mathbf{0}$, the system operates in a pure torque control mode using only the feedforward command, whereas setting $\boldsymbol{\tau}_{ff} = \mathbf{0}$ yields a conventional joint-space PD controller.

The command references are updated at the 1~kHz host rate, albeit the drives execute the control law \eqref{eq:copley} internally at 4~kHz. Torque tracking is achieved through a high-bandwidth inner current control loop operating at 16~kHz, where the commanded torque is converted to a quadrature-axis ($q$-axis) current reference using the motor torque constant relation $\mathbf{i}_q = \boldsymbol{\tau}_{cmd} \oslash \mathbf{k}_t$ with $\mathbf{k}_t$ denoting the vector of motor torque constants. Here, $\oslash$ represents element-wise division. Because the actuators operate within their nominal speed range for the target tasks, field weakening is not employed. The same relation is used in reverse to estimate motor torque from the measured current. These torque estimates, together with the measured joint position and velocity, are transmitted to the host within each control cycle. The actuator and transmission design, including low friction and low cogging characteristics enabled by the quasi-direct drive architecture, allows sufficiently accurate torque estimation without dedicated torque sensors, avoiding additional reflected inertia and mechanical complexity.

\textbf{Core Runtime Layer:} On the host computer, a Core Runtime Layer executes at 1~kHz independently of the selected control strategy. This layer ensures deterministic communication with the motor drives and provides the infrastructure required for deploying higher-level controllers. It consists of three subcomponents: Safety, Conversion, and Driver.

The Safety Layer enforces hardware limits by saturating commanded positions, velocities, and torques before transmission to the motor drives. The Conversion Layer accounts for mechanism-specific actuation effects that are not directly represented in the controller abstraction, including mechanically coupled joints and parallel linkages. For the torso joint, which is driven by two mechanically coupled motors, a leader–follower configuration is used, where the follower tracks an adaptive time-windowed, filtered estimate of the leader torque to prevent actuator contention. For the wrist's parallel linkages mechanism, precomputed lookup tables are used for command transformation. The complete conversion process requires approximately 3~$\mu$s, avoiding host-side bottlenecks and preserving the integrity of the 1~kHz control loop. Finally, the Driver Layer converts controller outputs into motor-drive-compatible units and performs EtherCAT data packaging and feedback decoding within each control cycle.

\textbf{Modular Control Layer:} On top of the Core Runtime Layer, a Modular Control Layer executes within the same 1~kHz real-time loop. This layer enables deterministic composition of task-specific control blocks that operate synchronously with the hardware interface while allowing different control strategies to be deployed depending on task requirements. The architecture was designed to support interchangeable control formulations without modifying the underlying communication or safety infrastructure.

For the baseball manipulation tasks presented in this work, control objectives were formulated either in joint space or in Cartesian (task) space, depending on task requirements. The joint-space controller follows a proportional–derivative structure with model-based feedforward compensation. Desired joint position and velocity references $\mathbf{q}_d, \dot{\mathbf{q}}_d \in \mathbb{R}^n$, together with gain vectors $\mathbf{k}_p, \mathbf{k}_d \in \mathbb{R}^n_{\geq 0}$, are commanded at the 1~kHz rate. Feedforward torque is computed using inverse dynamics based on the Recursive Newton–Euler Algorithm $\boldsymbol{\tau}_{ff} = \text{RNEA}(\mathbf{q}_d, \dot{\mathbf{q}}_d, \ddot{\mathbf{q}}_d)$, which compensates for inertial, Coriolis, and gravitational effects and enables the high-acceleration motions required for the target tasks. Accurate computation of inverse dynamics $\boldsymbol{\tau}_{ff}$ and forward kinematics relies on a calibrated robot model parameterized by both kinematic and dynamic properties. Kinematic parameters, including the full joint screw axis representation, were calibrated using the method in \cite{li2016} with measurements from OptiTrack markers attached to the end-effector frame. Dynamic parameters comprising link mass–inertial properties and joint Coulomb friction were identified using the geometric regularization approach in \cite{lee2020geometric} to mitigate parameter estimation bias from structural model mismatch, particularly passive forces induced by Bowden cable and wiring bending across joints. The resulting model yields a well-conditioned, positive-definite mass matrix.

The second control formulation is based on an Operational Space Controller (OSC) \cite{khatib1985_osc}, where motion objectives are specified in task space through desired end-effector position, velocity, and acceleration $\mathbf{x}_d, \dot{\mathbf{x}}_d, \ddot{\mathbf{x}}_d$. The resulting Cartesian impedance is mapped to joint torques using the manipulator Jacobian and dynamically consistent transformations. In this mode, joint-level proportional and derivative gains are set to zero, and the motor drives operate in pure torque control using the feedforward command channel. Because the arm is kinematically redundant, a null-space controller regulates internal posture without disturbing the primary end-effector objective. Despite the low reflected inertia of the quasi-direct-drive actuators—which can impose passivity-related limits on achievable stiffness \cite{colgate1997passivity}—the combination of low communication latency and high-bandwidth actuation {\color{editcolor2}allowed us to implement sufficiently high gains to enable seamless transitions between highly compliant interaction and stiff, high-speed free-space motion.} Joint stiffness values up to 400~Nm/rad were achieved for the large shoulder joints (J1–J4), 240~Nm/rad for the elbow roll (J5), and approximately 190 and 100~Nm/rad for the wrist joints (J6–J7). In operational space, Cartesian stiffness values up to 300~N/m with near-critical damping were obtained. These capabilities enable seamless transitions between highly compliant interaction and stiff, high-speed free-space motion within the same control framework.

 \textbf{Planning and Orchestration Layer :} Above the real-time control layers, a Planning and Orchestration Layer enables lower-frequency processes to operate outside the deterministic 1~kHz loop.
This layer generates reference parameters and coordinates controller behavior without imposing additional real-time constraints. Typical outputs include motion references like waypoints or spline parameters, controller configuration updates, and state-dependent switching of control blocks, which are then consumed by the 1~kHz control layer. Examples of this interaction between planning and control appear in the baseball manipulation tasks described in the subsequent sections.
}
{\color{editcolor2}
\subsubsection*{Torque Estimation}
\label{sec:torque_estimation}

The \athenazero{} hardware architecture relies on the central premise that output torques can be reliably estimated via the quadrature-axis current ($I_q$) measured by the motor controller. We emphasize that this methodology is not inherently novel; rather, we are directly leveraging a well-established and empirically proven paradigm within the broader robotics community, and albeit not perfect, can provide relatively accurate sensing with proper characterization. The reliability of current-based torque estimation in quasi-direct drive (QDD) systems has been extensively validated in dynamic legged locomotion, where high-bandwidth, proprioceptive torque control is critical. Foundational works on platforms such as the MIT Cheetah and Mini Cheetah \cite{wensing2017_cheetah, katz2016, seok2012_cheetah1}, as well as the Ghost Minitaur \cite{kenneally2016_minitaur}, explicitly demonstrate that accurate torque output and impact mitigation can be robustly achieved through phase currents. By adopting this approach, we anchor our specific hardware implementation within this community-wide consensus.

Technically, this framework depends on isolating the torque-producing current component through the standard Clarke and Park transformations of the motor's phase currents. In our current implementation, we assume standard field-oriented control where the direct-axis current is regulated to zero. Although we do not actively employ field weakening in these specific experiments, the control architecture maintains the option to utilize it for high-speed operations. The fundamental relationship linking our controller to the physical world is governed by the standard ideal torque equation, $\tau = K_t I_q$ where $\tau$ is the output torque and $K_t$ is the motor's torque constant.

\begin{figure}
\centering
\includegraphics[width=0.9\textwidth]{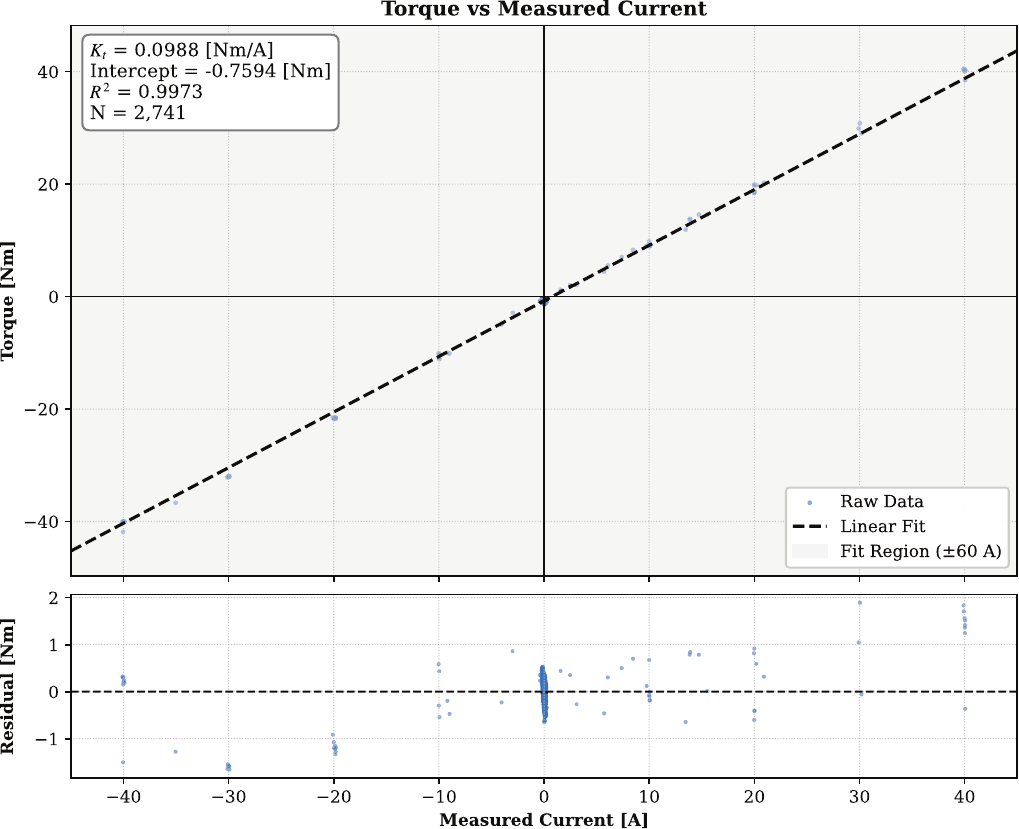}
\caption{\textcolor{editcolor}{\textbf{Dynamometer characterization of output torque in linear regime.} Leveraging a dynamometer, we characterize the torque production capabilities of a QDD actuator. During a locked rotor test, the system swepdf different commanded torques and measures the resultant torque through a load cell. The data are fit via least squares, and a highly linear model representation with an $R^2$=0.9973 is noted for 2,741 samples. }}
\label{fig:dyno}
\end{figure}

Although we have conducted extensive characterization of our motors across various dynamic regimes, the scope of this supplementary section is intentionally focused on providing a straightforward validation of our transmission's baseline efficiency. To this end, we rely on data gathered from a locked-rotor test (Fig.~\ref{fig:dyno}) on a custom dynamometer, similar to as described in \cite{lee2019empirical, nazon2025}. This empirical test serves as a practical baseline, yielding a highly linear fit between commanded current and measured output torque.

From this empirical fit, we extracted a measured torque constant ($K_t$) of 0.09888~Nm/A for a 76~mm frameless stator-rotor pair equipped with a 10:1 planetary gearbox. Correcting for the discrepancy with the manufacturer’s nominal specification of 0.115~Nm/A, our empirical characterizations routinely reveal that manufacturer-provided $K_t$ estimates are inflated by roughly 15--20\%. This experimental data, deliberately collected in the expectedly linear region of the actuator, confirms that mapping the $I_q$ current to output torque is indeed linear over the tested regime and yields sufficient accuracy for our hardware. Because our system utilizes a high-efficiency, low-ratio QDD transmission, the parasitic friction and substantial hysteresis that typically plague traditional high-ratio gearboxes are largely mitigated. Consequently, estimating torque directly via calibrated phase currents avoids the mechanical fragility that routinely degrades the performance of external strain-gauge sensors. Ultimately, this simplified characterization provides the necessary confidence for the kinodynamic claims made in the main text, though we recognize the boundaries of this validation. Factors such as actuator bandwidth, long-term thermal drift, and complex dynamics under high-acceleration regimes are critical to a holistic understanding of the system but fall beyond the scope of this current manuscript. We plan to detail our full dynamic characterization in future work.

}
\subsubsection*{Dynamic Manipulation Tasks}
\label{sec:task_def}

{\color{editcolor2}The low-inertia architecture of \athenazero{} necessitates the development of specialized control strategies, particularly for highly dynamic manipulation.} To empirically ground this exploration, we investigate one end of the dynamic spectrum through throwing, catching, and batting as primary case studies. These tasks not only demonstrate the unique capabilities afforded by the system's hardware, but also establish foundational primitives for future dynamic behaviors (Fig. \ref{fig:catch_throw_bat}). Fundamentally, these tasks represent complex problems of energy transfer, governing the precise mechanics of how and when mechanical energy is exchanged between the manipulator and the object.

\textbf{Throwing} a ball with high velocity and accuracy is an almost uniquely human ability. This act represents a feedforward energy transfer problem: the ball starts at a nominal resting point and accelerates while resisting the manipulator according to its mass. Consequently, accurate throwing relies heavily on an equally accurate internal model of the robot's dynamics. This task serves as a catalyst for studying dynamic trajectory generation in the presence of disturbances (such as varying ball mass) and underscores the need for advanced calibration techniques to achieve accurate internal modeling (see Throwing subsection).

\textbf{Catching} a ball at variable speeds is a feedback energy transfer problem under uncertainty. Formulating this simply as the inverse of throwing fails to capture the critical nuances between the two. Specifically, catching is better conceptualized as the robot's ability to define its desired state to control a future interaction. Although the ball follows its ballistic trajectory, the robot has a very short period of time (250–400~ms) to update its pose. It must move into position to manage large, potentially uncertain impulses while strictly respecting force, torque, and stability limits. Ultimately, this task demands the careful design of low-latency impedance feedback control to adapt to uncertainties in ballistic predictions and successfully secure the ball.

\textbf{Batting} addresses a historically difficult problem in robotics: control under closed-chain constraints. As two arms rigidly coordinate through a shared object, the system transitions from free-space independence to joint interdependence, creating the risk of large internal forces if constraints are not precisely managed. In batting, this challenge is compounded. The robot must generate and adapt full-body swing trajectories involving high accelerations and decelerations, all while preparing for impact with an incoming ball traveling in the opposite direction. This combination of closed-chain coordination, rapid adaptation, and impact dynamics requires developing online planning and control methods that extend well beyond the capabilities of standard industrial manipulators.

Collectively, these tasks highlight two fundamental challenges in robotic control. The first is the capacity for rapid acceleration and deceleration. Without achieving peak end-effector velocity at the precise kinematic phase, throwing and batting motions lose efficacy; similarly, without sufficient acceleration toward a predicted interception waypoint, a catch will fail. These stringent temporal constraints underscore the critical importance of millisecond-level precision in dynamic manipulation. The second challenge is the management of physical interaction forces. Catching necessitates the controlled dissipation of a projectile's kinetic energy upon impact. Batting further compounds this by requiring the system to absorb and redirect impulsive collision forces while simultaneously mitigating the internal antagonistic forces inherent to closed-kinematic-chain motions. By capturing these dual demands---generating high-bandwidth motion and sustaining robust physical interaction---these tasks serve as useful benchmarking primitives for dynamic manipulation research.

\subsubsection*{Capability: Throwing}
{\color{editcolor}
Dynamic throwing requires the control of a kinetic chain---a phenomena where velocities of sequential serial joints reach maximum velocity one-after-another, in order to efficiently transfer momentum and energy from the chain into the ball's release velocity. This method mirrors how a whip can break the sound barrier at its tip via a proximal input of force. Interestingly, throwing with accuracy and power is an almost uniquely human behavior, an anatomical adaptation believed to have evolved to improve hunting efficacy \cite{roach2012biomechanics}. Throughout the throwing motion energy is transferred from the ground, into the trunk of the body, into the shoulder, into the elbow, and finally, out through the hand. Release timing and arm slot play a vital role in the accuracy and speed of the throw.

In contrast to the highly dynamic nature of human throwing, robotic tossing and throwing have traditionally relied on fundamentally different paradigms \cite{zeng2020tossingbot}. Historically, robotic systems have not exploited their own rigid-body dynamics to efficiently transfer energy into a projectile; rather, they perform purely kinematic trajectories, swinging at a specified velocity and releasing the object at a computed waypoint. Consequently, these motions often lack fluidity, and mechanical energy is frequently dissipated through gearbox friction and parasitic vibration instead of being channeled into the throw. This limitation stems primarily from traditional hardware and control methodologies. Because conventional motion planners have been developed under a strict kinematic framework for decades, they do not easily scale to the complex, high-energy dynamical requirements of explosive tasks. However, as robotic hardware becomes increasingly capable of dynamic operation, the literature is beginning to reflect a shift. Although still in the minority, emerging research has demonstrated dynamic robotic throwing utilizing both model-based \cite{ishikawa2008throwing} and learning-based \cite{fey2025bridgingsimtorealgapathletic} methodologies.

The objective of throwing is to induce a dynamic whipping motion within the manipulator, efficiently transferring energy proximally from the base to the distal end-effector. To systematically investigate parametric variations in the robot's dynamics, this preliminary study employs a model-based framework driven by dynamical trajectory optimization, specifically formulated via direct collocation.

Whole-arm trajectories were generated using SNOPT\cite{gill2005snopt} by iteratively applying costs and constraints on the computed cubic splines and bootstrapping from prior solutions. The optimization objective was to regulate joint accelerations and torques while satisfying the terminal constraint of a target end-effector release velocity vector. The resulting trajectories exhibited pronounced dynamic behavior, effectively exploiting the system's kinetic chain, particularly at target release velocities of $\geq$20~m/s.
}

During physical execution of the computed trajectories, two notable sim-to-real discrepancies emerged. First, the dynamics of ball release and roll-off proved highly challenging to model. The latency between the initial hand-opening command and the physical detachment of the projectile varied between 39 and 67 ms, highly dependent on the projectile's mass and the instantaneous state of the manipulator (position, velocity, and acceleration). Because conventional tactile sensors lacked the temporal resolution to measure this variance, a high-speed camera operating at 1000 frames per second was utilized for empirical observation. This analysis facilitated the derivation of robust roll-off timing models for both tennis balls and baseballs across a range of velocities, ultimately yielding highly accurate targeting (within a 25 cm x 25 cm bounding box) at distances of $\ge$5~m. The second discrepancy involved trajectory tracking errors stemming from unmodeled hardware dynamics and the accelerations profiles commanded to our system. Due to the explosive nature of the throwing motion, the motor controllers exhibited imperfect torque tracking, exacerbated by non-linear mechanical phenomena such as cogging, friction, and structural vibration. Appropriately-tuned tracking was achieved by applying high velocity gains and low positional gains, augmented by feedforward torques computed via the inverse dynamics of the reference trajectory.

To mitigate the affect of unmodeled dynamics and system uncertainty, we implemented a near-online trajectory adaptation framework. For a designated target, the manipulator executed an initial throw based on a nominal release velocity. Utilizing the empirical error of the projectile's impact location, a secondary nonlinear kinematic trajectory optimization procedure locally refined the underlying dynamic trajectory by updating the release vector or release timing parameters. Because this optimization converged in approximately 3 seconds, the system could perform rapid, sequential updates, efficiently minimizing the targeting error over successive iterations.

\subsubsection*{Capability: Catching}
Whereas throwing necessitates the efficient transfer of kinetic energy to a projectile, catching intrinsically relies on optimal energy dissipation. When humans catch, they are known to dynamically adjust their ``task-readiness impedance," setting the stiffness of their arms to better absorb the ball’s motion prior to contact \cite{tanaka2003task}. This approach has generally been unavailable to catching robots due to the high reflected inertia of standard manipulator hardware, which fundamentally limits the ability to impedance-match lightweight objects. As a result, previous robotic catching frameworks have primarily depended on alternative kinematic or mechanical strategies to attenuate impact forces.

The most common of these strategies is to plan trajectories that align the end-effector velocity with that of the incoming object~\cite{hove1991experiments, mirrazavi2016dynamical}. Albeit successful for catching relatively slow-moving objects, a pure velocity-matching approach will have limited robustness as ball speeds increase, especially when velocity estimates are noisy or uncertain. Moreover, the extensibility of velocity-matching approaches to static-grasping tasks is low as it would require the robot to cumbersomely come to a full stop prior to making contact. Another method seeks to minimize impulse by planning contacts perpendicular to the ball’s velocity~\cite{yan2024impact}. Though this has shown success in dual-arm catching systems, it is less directly applicable to human-like single-arm catching behaviors. The low reflected inertia design of \athenazero{} makes it uniquely suited for an impedance-matching approach, with the arm exhibiting effective mass similar to the weight of a baseball in many catching configurations (see Fig.~\ref{fig:impedance}).

Our catching behavior tracks the motion of a baseball with reflective tape using an OptiTrack motion capture system. An inverse kinematic problem is then solved to produce a catching configuration that intercepts the forward-predicted ballistic trajectory of the ball. Although \athenazero{} was designed to minimize reflected actuator inertia and concentrate mass proximally to achieve a low end-point mass, the effective mass during impact also depends on the robot’s configuration. Thus, selecting an appropriate posture within the null space of the end-effector positioning task can further enhance impedance matching. To this end, costs in the IK solver push solutions toward offline-generated seeds that penalize high effective mass configurations. A simple, computationally efficient motion planner then produces quintic joint trajectories---executed using an identified dynamic model for the manipulator coupled with high gains to maintain high speed and precision---to bring the robot into the catch configuration. Re-planning is performed continuously as the perception module updates the predicted intercept location. Shortly before the anticipated intercept time, the arm transitions from high-gain joint-space control to an operational-space controller that renders anisotropic stiffness in task space. The manipulator remains stiff in directions orthogonal to impact to enable visual-servoing if new intercept estimates are received during this phase, but behaves like a soft spring in the impact direction to help absorb the ball's energy.

\subsubsection*{Capability: Batting}
Batting, similar to throwing, is an energy-transfer process that leverages the kinetic chain to maximize the energy imparted to a flying ball. Unlike throwing, batting engages both arms, enabling stronger and more precise control of the bat by distributing its weight and the impact forces across both limbs. However, gripping the bat with two hands creates a closed-chain system, where kinematic constraints can induce internal forces if the arms are not well coordinated. Classically \cite{tsai1999robot}, this demands on careful planning and control, a challenge further amplified in batting by the motion’s highly dynamic nature.
Batting is, on the other hand, similar to catching in that it requires rapid adjustment of motion to the ball’s trajectory, coupled with readiness to withstand the ensuing impact. Unlike catching, however, the objective is not to dissipate the ball’s energy and bring it to rest, but to impart energy through the bat to redirect its flight; an interaction that produces a more forceful impact.

Placing the burden of handling such dynamic, impact-tolerant motion in an overconstrained system solely on planning would require accurate prediction of the task over its full time horizon---an approach that is not only inefficient but also heavily dependent on precise models of the involved systems and their interaction dynamics. Augmenting this with feedback control, for instance by applying corrections based on sensed wrenches \cite{kumar1988force}, is likewise impractical given the bandwidth limitations of sensing and actuation in such fast dynamics. These challenges, however, can be alleviated by exploiting the unique physical characteristics of \athenazero{}: with appropriate tuning of joint compliance and damping, its backdrivability can passively mitigate residual internal forces arising from unplanned deviations in closed-chain motion, as its inherently low effective mass contributes to safer and more controlled impact behavior with the flying ball.

Similar to the throwing task, the problem is formulated as an optimization to maximize the bat’s velocity at a given strike position, solved using SNOPT \cite{gill2005snopt}. The formulation, however, differs in two important ways. First, an accelerated optimization pipeline is employed that incorporates the closed-chain constraint, allowing only partial orientation alignment between the two hands. More importantly, the optimized trajectories are represented in the phase domain, which decouples their geometric characteristics from their temporal evolution \cite{shahriari2024path}. This representation enables smooth blending between neighboring optimized trajectories when the estimated strike position changes, even if the trajectories differ in timing, such as overall duration. Furthermore, it allows the execution of different motion types (such as forward, reverse, cyclic) simply by adjusting the phase trajectory, making the process more intuitive. The control strategy followed the same structure as in the other tasks, with the key difference of training a model to generate residual corrective motions that were superimposed on the optimized trajectory \cite{zeng2020tossingbot}, compensating for disturbances in tracking performance caused, for instance, by unmodeled bat dynamics.

\subsection*{Supplementary Movie Captions}

\textbf{Movie s1: }
Movie s1 illustrates the inherent backdrivability of the \athenazero{} robot. The first two scenes depict a human interacting with the robot while motors are powered off. As forces are imparted via external perturbations from the human, energy is absorbed and dissipated through the entire kinematic chain of the robot. This intuitively underscores the low inertia properties of the robot. The final scene in this movie illustrates a scenario where the robot is powered on and is commanded to adjust stiffness parameters while interacting with an external force.
\\
\noindent \textbf{Movie s2: }
Movie s2 showcases the impact tests performed on the Franka FR3 and \athenazero{}. Notably, the low inertia properties of \athenazero{} allow the system to better absorb impact as compared to arms with higher gear ratios.
\\
\noindent \textbf{Movie s3: }
Movie s3 showcases a scenario where two \athenazero{} robots participate in a game of catch.
\\
\noindent \textbf{Movie s4: }
Movie s4 showcases a scenario where two \athenazero{} robots participate in batting practice.
\\
\noindent \textbf{Movie s5: }
Movie s5 showcases the tactile sensing capabilities of the \athenazero{} robot. All motions of the robot were commanded via teloperation and a visualization of the tactile array is provided on-screen.
\\

\end{document}